%% file: main.tex
\documentclass[letterpaper,11pt]{article}
\usepackage{etex}
\usepackage[margin=1in]{geometry}

\usepackage[dvipsnames,table,xcdraw]{xcolor}
\usepackage{amssymb,amsfonts,amsmath,amstext,amsthm,mathrsfs}
\usepackage{graphics,latexsym,epsfig,psfrag,wrapfig,comment,paralist}
\usepackage{xspace}
\usepackage{subcaption,bm,multirow}
\usepackage{booktabs}
\usepackage{mathdots}
\usepackage{bbold}
\usepackage{centernot}
\usepackage{algorithm}
\usepackage{algorithmic}
\usepackage{prettyref}
\usepackage{listings}
\usepackage{tikz}
\usetikzlibrary{decorations,calligraphy}
\usepackage{pgfplots}
\usetikzlibrary{decorations.pathreplacing}
\usetikzlibrary{shapes}
\usetikzlibrary{positioning, arrows, matrix, calc, patterns, fit}
\usepackage{caption}
\usepackage{array}
\usepackage{mdwmath}
\usepackage{multirow}
\usepackage{mdwtab}
\usepackage{eqparbox}
\usepackage{multicol}
\usepackage{amsfonts}
\usepackage{multirow,bigstrut,threeparttable}
\usepackage{amsthm}
\usepackage{array}
\usepackage{bbm}
\usepackage{epstopdf}
\usepackage{mdwmath}
\usepackage{mdwtab}
\usepackage{eqparbox}
\usepackage{tikz}
\usepackage{latexsym}
\usepackage{amssymb}
\usepackage{bm}
\usepackage{graphicx}
\usepackage{mathrsfs}
\usepackage{epsfig}
\usepackage{psfrag}
\usepackage{setspace}
\definecolor{linkColor}{HTML}{E74C3C}
\definecolor{pearcomp}{HTML}{B97E29}
\definecolor{citeColor}{HTML}{2980B9}
\definecolor{urlColor}{HTML}{1D2DEC}
\definecolor{conjColor}{HTML}{9ab569}
\usepackage[CJKbookmarks=true,
            bookmarksnumbered=true,
            bookmarksopen=true,
            colorlinks=true,
            citecolor=citeColor,
            linkcolor=linkColor,
            anchorcolor=red,
            urlcolor=urlColor,
            ]{hyperref}
\usepackage{natbib}
\setcitestyle{authoryear,round}
\newtheoremstyle{break}
  {\topsep}{\topsep}%
  {\itshape}{}%
  {\bfseries}{}%
  {\newline}{}%

\usepackage{pgfplots}
\usepackage{amsmath,amssymb}
\usepackage{mathtools}
\usepackage{stmaryrd}
\pgfmathdeclarefunction{gauss}{3}{%
  \pgfmathparse{#3/(#2*sqrt(2*pi))*exp(-((x-#1)^2)/(2*#2^2))}%
}
\pgfmathdeclarefunction{mingauss}{4}{%
  \pgfmathparse{#4/(#3*sqrt(2*pi))*exp(-max((x-#1)^2, (x-#2)^2)/(2*#3^2))}%
}

\tikzset{
  invisible/.style={opacity=0},
  visible on/.style={alt={#1{}{invisible}}},
  alt/.code args={<#1>#2#3}{%
    \alt<#1>{\pgfkeysalso{#2}}{\pgfkeysalso{#3}}
  },
}

\newtheorem{definition}{\textbf{Definition}}%

\newtheorem{lemma}{\textbf{Lemma}}%
\newtheorem{theorem}{\textbf{Theorem}}%

\newtheorem{assumption}{Assumption}

\newtheorem*{lemmai*}{\textbf{Lemma (informal)}}
\newtheorem{remark}{\textbf{Remark}}

\usepackage[normalem]{ulem}
\usepackage[nameinlink]{cleveref}

\crefname{assumption}{assumption}{assumptions}
\crefname{table}{table}{tables}
\crefname{claim}{claim}{claims}

\renewcommand{\cite}[1]{\citep{#1}}

\newcommand{\1}{\mathbf{1}}

\usepackage[utf8]{inputenc} %
\usepackage[T1]{fontenc}    %
\usepackage{hyperref}       %
\usepackage{url}            %
\usepackage{booktabs}       %
\usepackage{amsfonts}       %
\usepackage{nicefrac}       %
\usepackage{microtype}      %
\usepackage{xcolor}         %

\input{math_commands}

\usepackage[framemethod=tikz,%
innerleftmargin=\parindent,%
skipabove=0.6\baselineskip,%
skipbelow=0.6\baselineskip,%
innertopmargin=0.4\baselineskip,%
innerbottommargin=0.4\baselineskip]{mdframed}

\newlength{\defparindent}
\newmdenv[linewidth=0.4pt,%
linecolor=black,%
backgroundcolor=white,%
settings={\setlength{\parindent}{\defparindent}}]{singleframed}

\usepackage{mdframed}

\usepackage{amsmath}
\usepackage{amsthm}
\usepackage{hyperref}
\usepackage{url}
\usepackage{cleveref}
\usepackage{amssymb}

\usepackage{graphicx}
\usepackage{subcaption} %

\usepackage{color}
\definecolor{cm}{RGB}{0,0,200}
\definecolor{purple}{RGB}{200,0,200}

\usepackage[intoc]{nomencl}
\makenomenclature

\makeatletter
\newcommand{\vast}{\bBigg@{2.5}}
\newcommand{\Vast}{\bBigg@{5}}
\makeatother

\usepackage{bbm}

\usepackage[utf8]{inputenc}
\usepackage[LGR,T1]{fontenc}

\usepackage{breqn}

\usepackage{enumitem,kantlipsum}
\newcommand\blfootnote[1]{%
  \begingroup
  \renewcommand\thefootnote{}\footnote{#1}%
  \addtocounter{footnote}{-1}%
  \endgroup
}
\title{Efficient Prompt Caching via Embedding Similarity}

\author{Hanlin Zhu $^\dagger$\quad
Banghua Zhu $^\dagger$ \quad
Jiantao Jiao $^{\dagger, \ddagger}$
 \blfootnote{Emails: \texttt{\{hanlinzhu,banghua,jiantao\}@berkeley.edu} }\\ {}\\ 
$^\dagger$Department of Electrical Engineering and Computer Sciences, UC Berkeley\\
$^\ddagger$Department of Statistics, UC Berkeley\\
 { } \\
}
\date{\today}
\pgfplotsset{compat=1.18} 
\begin{document}

\maketitle

\begin{abstract}
Large language models (LLMs) have achieved huge success in numerous natural language process (NLP) tasks. However, it faces the challenge of significant resource consumption during inference. In this paper, we aim to improve the inference efficiency of LLMs by prompt caching, i.e., if the current prompt can be answered by the same response of a previous prompt, one can directly utilize that previous response without calling the LLM. Specifically, we focus on the prediction accuracy of prompt caching for single-round question-answering tasks via embedding similarity. The existing embeddings of prompts mostly focus on whether two prompts are semantically similar, which is not necessarily equivalent to whether the same response can answer them. Therefore, we propose a distillation-based method to fine-tune the existing embeddings for better caching prediction. Theoretically, we provide finite-sample guarantees for the convergence of our method under different types of loss functions. Empirically, we carefully construct a hard dataset based on \citet{47761} where the existing embedding model \citep{wang2022text} only achieves an AUC of 0.51. We then fine-tune the above embedding model, which significantly improves the AUC of caching prediction from 0.51 to 0.81. 
We also conduct simulations demonstrating that our trained models achieve better caching efficiency than the previous embedding model.
\end{abstract}

\input{intro}
\input{prelim}
\input{theory}
\input{exp}

\input{conclusion}

\bibliographystyle{plainnat}
\bibliography{references}

\newpage
\appendix

\input{appendix}

\end{document}

%% file: math_commands.tex
\usepackage{amsmath,amsfonts,bm}

\def\1{\bm{1}}

\newcommand{\E}{\mathbb{E}}

\DeclareMathAlphabet{\mathsfit}{\encodingdefault}{\sfdefault}{m}{sl}
\SetMathAlphabet{\mathsfit}{bold}{\encodingdefault}{\sfdefault}{bx}{n}

\newcommand{\sigmoid}{\sigma}

\DeclareMathOperator*{\argmin}{arg\,min}

\PassOptionsToPackage{table}{xcolor}
\usepackage{tikz}
\usetikzlibrary{positioning, shapes.geometric, arrows.meta}

\definecolor{lightpink}{rgb}{1, 0.85, 0.9}
\definecolor{lightblue}{rgb}{0.68, 0.85, 0.9}
\definecolor{lightyellow}{rgb}{1.0, 1.0, 0.88}

%% file: intro.tex
\section{Introductions}
\label{sec:intro}

The recent development of large language models (LLMs) and foundation models has notably enhanced the potential of AI systems~\citep{ziegler2019fine,wei2022emergent,chowdhery2022palm,ouyang2022training,bubeck2023sparks,nori2023capabilities,openai2023gpt4,beeching2023stackllama,anil2023palm}. However, due to the large scale of those models, it causes significant resource consumptions not only during the training process, but also in the inference stage~\citep{sharir2020cost,patterson2021carbon,bommasani2022opportunities}. Moreover, the latency of LLMs during inference is not negligible since the model only generates one token at a time due to its auto-regressive nature, which makes it unfavorable to be applied to systems desiring high throughput, such as search engines~\citep{zhu2023optimal}. Therefore, it would be appealing to reduce the resource consumption and latency without degrading the performance of LLMs.  
\begin{figure}[htbp]
    \centering
    \includegraphics[width=0.7\textwidth]{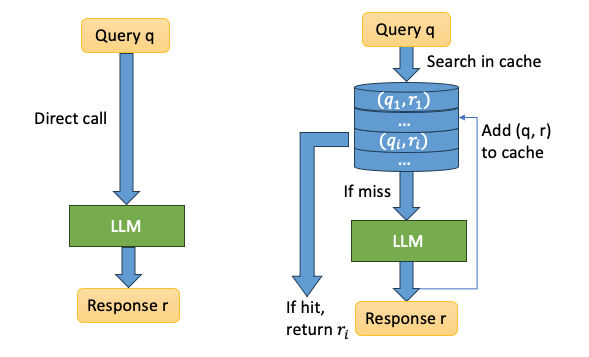}
    \caption{The procedure of calling LLMs with or without cache. When a cache is available, one can store some of the previous prompt-response pairs in the cache, and for a new prompt, one can search in the cache whether a prompt and the current prompt can be answered by the same response. If there is a hit, one can directly reuse the response of the previous prompt without calling LLMs.}
    \label{fig:caching}
\end{figure}

A natural idea to reduce resource consumption and latency is to reduce the number of calls to LLMs, which can be implemented by caching, a technique that has a long history of being studied and applied to important areas such as computer architecture and web retrieval ~\citep{smith1982cache,wang1999survey,kumar2016overview}. \citet{zhu2023optimal} studies prompt (or query) caching for LLMs, i.e., some of the previous prompt-response pairs are stored in a cache with limited size, and whenever a new prompt arrives, one can search in the cache whether a prompt has the same semantic meaning as the current prompt, and can directly reuse the response of the previous prompt without calling LLMs if there is a hit (see \Cref{fig:caching} for a figurative illustration).

 \citet{zhu2023optimal} focuses on caching algorithm design and directly assumes a semantic search oracle. Although previous literature studies semantic search or embedding-based methods~\citep{bast2016semantic,chang2020pre,kamalloo2023evaluating}, which could serve as solutions to the caching hit problem~\citep{zilliztech_2023}~\footnote{i.e., searching whether there exists a prompt in the cache such that the current prompt can be answered by the same response.}, it is challenging to obtain a good embedding that can accurately represent the semantic meaning of a prompt. Moreover, a semantically similar prompt pair cannot necessarily be answered by the same response, which implies that we need a different vector embedding specifically for the caching hitting problem that can be used to search a similar prompt more efficiently and better predict whether a pair of prompts can be answered by the same response.

\begin{figure}[htbp]
    \centering
    \subfloat[A query $q$ can be embedded into a $d$-dimensional vector $v_\theta(q)$, where $v_\theta(\cdot)$ can be the last layer of another language model parameterized by $\theta$.]{
        \includegraphics[width=0.4\textwidth]{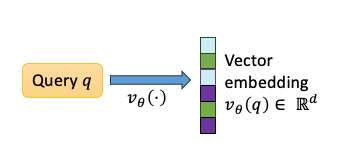}
        \label{fig:caching_embedding}
    }
    \hfill
    \subfloat[To determine whether two prompts $q_1$ and $q_2$ can be answered by the same response, we calculate the cosine similarity between the correspond embeddings $v_\theta(q_1)$ and $v_\theta(q_2)$. When the cosine similarity is below a given threshold, they can be answered by the same response (a hit); otherwise, they need to be answered by different responses (a miss).]{
        \includegraphics[width=0.4\textwidth]{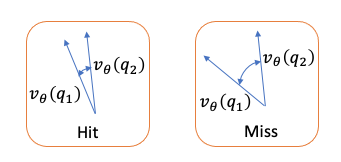}
        \label{fig:cos_sim}
    }
    \caption{\Cref{fig:caching_embedding} shows the embedding procedure. \Cref{fig:cos_sim} illustrates how to utilize the cosine similarity to determine whether two prompts can be answered by the same response.}
    \label{fig:embedding_caching_hit}
\end{figure}

In this paper, we aim to learn a good vector embedding such that the similarity of embeddings of a prompt pair could better encode the information of whether the pair of prompts can be answered by the same response (see \Cref{fig:embedding_caching_hit}). We propose a distillation-based method, which aims to learn whether a prompt pair can be answered by the same response via cosine similarity of the embeddings of the prompt pair, to fine-tune an existing semantic vector embedding from~\citet{wang2022text}. Theoretically, we provide finite sample guarantees for the learning error under mild assumptions using two different loss functions: binary cross entropy (BCE) and squared log difference (SLD), respectively (\Cref{sec:theory}). Empirically, we carefully construct a hard dataset based on \citet{47761} (\Cref{sec:exp_construct_dataset}) and fine-tune the embedding from \citet{wang2022text} , which significantly improves the AUC of caching prediction from 0.51 to 0.81 (\Cref{sec:exp_fine_tuning}). We also conduct simulations of the prompt streaming with caching using different embeddings, which demonstrates that our trained embedding models achieve better caching efficiency than the previous one. (\Cref{sec:exp_caching_sim}).

\subsection{Related works}
\label{sec:related_work}

\paragraph{Caching.} Caching algorithms are important to computer architecture and systems and have long been explored~\citep{lee2001lrfu,stallings2011operating,bura2021learning}. In recent years, caching has also been applied to online learning analysis and machine learning advice~\citep{he2017deep,chang2018learn,jiang2019multi,shuja2021applying,mukhopadhyay2021online,faizal2022regret}. \citet{bang2023gptcache} builds the framework for caching the responses from LLMs. \citet{zhu2023optimal}  studies optimal caching algorithm for prompt in both online and offline learning settings. Instead of studying caching policy, we aim to study how to efficiently search and accurately predict whether there is a caching hit.

\paragraph{Retrieval-based LLMs.} A line of work studies augmenting a language model by retrieval-based method~\citep{grave2016improving,grave2017unbounded,khandelwal2019generalization,borgeaud2022improving,izacard2022few,zhong2022training,min2022nonparametric}. For example, the kNN-LM model~\citep{khandelwal2019generalization} interpolates a distribution obtained by the vector embedding of $k$ nearest neighbors with the distribution of language models. Our formulation of probability via embedding similarity is inspired by these works.

%% file: prelim.tex
\section{Preliminaries}
\label{sec:prelim}

We introduce in this section some basic notations, definitions, and assumptions. Let $\mathcal{Q}$ denote the set of all possible prompts (queries). For any prompt pair $(q_1, q_2) \in \mathcal{Q} \times \mathcal{Q}$, we denote the ground-truth probability that $(q_1, q_2)$ can be answered by the same response by $P^\star(q_1=q_2)$. Rigorously speaking, the ground-truth probability should be either 1 or 0, since for any two queries, we shall exactly know whether they can be answered by the same response. However, although the ground truth is always deterministic, we can still train a probabilistic predictor whose output represents the confidence of the prediction. On the other hand, even if a prompt pair can be answered by the same response in ground truth, the label (either by a human or a language model) might still be noisy at different levels. Therefore, we allow the ground-truth probability to be any real number in $[0,1]$.

Assume there exists an underlying distribution $\mu$ over prompt pairs $(q_1, q_2)$. 
Note that we do not have direct access to the $\mu$ and instead are given a dataset 
\[
\mathcal{\mathcal{D}} = \{ (q_{i,1}, q_{i, 2}, p_i)\}_{i=1}^N,
\]
 where $(q_{i,1}, q_{i,2}) \stackrel{\text{i.i.d.}}{\sim} \mu$ and $p_i \in [0, 1]$ with 
 $p_i \sim \mathcal{P}(\cdot| q_{i,1}, q_{i,2})$~\footnote{$\mathcal{P}(\cdot| q_{i,1}, q_{i,2}) \in \Delta([0,1])$ can be any distribution on $[0,1]$.} and 
 \[\mathbb{E}[p_i | q_{i,1}, q_{i,2}] = P^\star(q_{i,1}=q_{i,2}).\] Below, we define the vector embedding of prompts (\Cref{defn:embedding}) and probability via embedding similarity (\Cref{defn:induced_prob}).

\begin{definition}[Embedding of prompts]
\label{defn:embedding}
For any prompt $q$, let $v_\theta(q) \in \mathbb{R}^d$ denote its vector embedding (see \Cref{fig:caching_embedding}) where $v$ can be viewed as the mapping of prompts to a specific layer of a language model, and $\theta \in \Theta$ is the parameters of that model.
\end{definition}

\begin{definition}[Probability via embedding similarity]
\label{defn:induced_prob}
For any two prompts $q_1, q_2$, we denote the induced probability via embedding similarity that $q_1,q_2$ can be answered by the same response by (see \Cref{fig:cos_sim})
\begin{align*}
    P_{\theta,\lambda,c}(q_1 = q_2) \triangleq \sigmoid(\texttt{sim} ( v_\theta(q_1), v_\theta(q_2)) / \lambda - c),
\end{align*}
where $\texttt{sim}(\cdot, \cdot)$ denotes the cosine similarity, i.e., \[\texttt{sim}(x, y) = \frac{\langle x, y \rangle}{\|x\| \cdot \|y\|}\] for two vectors $x, y \in \mathbb{R}^d$, $\sigmoid(\cdot)$ denote the sigmoid function, i.e.,
\[\sigmoid(x) = \frac{1}{1+\exp(-x)}\] for $x \in \mathbb{R}$, and $\lambda \in \Lambda \subset \mathbb{R}_+, c \in \mathcal{C} \subset \mathbb{R}$ are two real-valued parameters.

\end{definition}

We also make the following realizability and boundedness assumptions for theoretical analysis.

\begin{assumption} [Realizability]
\label{assump:realizability}
Assume there exists $\theta^\star \in \Theta, \lambda^\star \in \Lambda, c^\star \in \mathcal{C}$, s.t. for any prompt pairs $(q_1, q_2) \in \mathcal{Q} \times \mathcal{Q}$, it holds that 
\begin{align*}
    P_{\theta^\star, \lambda^\star, c^\star}(q_1 = q_2) = P^\star(q_1 = q_2).  
\end{align*}
\end{assumption}

\begin{assumption}[Boundedness]
\label{assump:boundedness}
    Assume there exist constants $L_\lambda, B_c > 0$, s.t. 
    \begin{align*}
         \lambda \geq L_\lambda, |c| \leq B_c, \ \ \forall \lambda \in \Lambda, c \in \mathcal{C}.
    \end{align*}
\end{assumption}

\paragraph{Additional notations.} For convenience, for any $p \in [0,1]$, we denote $\bar p = 1 - p$. Also, for any prompt pairs $q_1, q_2$, we denote \[\bar P^\star(q_1 = q_2) = 1 - P^\star(q_1=q_2)\] and \[\bar P_{\theta,\lambda,c}(q_1 = q_2) = 1 - P_{\theta,\lambda,c}(q_1=q_2).\] 
We also use $\| \cdot \|_{2,\mu}$ to denote the $(2,\mu)$-norm given a distribution $\mu$, where for any function $f$ with input $x$, $\| f \|_{2,\mu}^2 = \mathbb{E}_{x \sim \mu}[f^2(x)]$. We also use $\texttt{clip}(x, a, b) = \max(a, \min(x, b))$ to clip $x$ to keep the value in the interval $[a, b]$.

%% file: theory.tex
\section{Algorithms and Theoretical Results}
\label{sec:theory}

In this section, we introduce our main algorithms for fine-tuning embeddings using labels distilled from GPT-4~\citep{openai2023gpt4}, and provide finite sample guarantees for the convergence of the learning error. We compare two different loss functions, i.e., binary cross entropy loss (\Cref{sec:BCE}) and squared log difference loss (\Cref{sec:squared_log_diff}).

\subsection{Algorithms for embedding fine-tuning}
\label{sec:alg_fine_tuning}

We first present our main algorithm for embedding fine-tuning in \Cref{alg:fine_tuning_theoretical}.

\begin{algorithm}[h]
\caption{Supervised embedding fine-tuning (theoretical version)}
\label{alg:fine_tuning_theoretical}
\begin{algorithmic}[1]
\STATE \textbf{Input:} Dataset $\mathcal{\mathcal{D}} = \{ (q_{i,1}, q_{i, 2}, p_i)\}_{i=1}^N$, parameter spaces $\Lambda, \mathcal{C}, \Theta$. loss type $L$.
\IF{$L$ = BCE}
\STATE  $\mathcal{L}_{\mathcal{D}} \gets \mathcal{L}^{\text{BCE}}_{\mathcal{D}}$ as defined in \eqref{eq:BCE_loss_empirical}
\ELSIF{$L$ = SLD}
\STATE $\mathcal{L}_{\mathcal{D}} \gets \mathcal{L}^{\text{SLD}}_{\mathcal{D}}$ as defined in \eqref{eq:sld_loss_empirical}
\ELSE
\STATE Raise Not Implemented Error
\ENDIF
\STATE  $\hat \theta, \hat\lambda, \hat c \gets \arg\min_{\theta\in\Theta, \lambda\in\Lambda, c\in\mathcal{C}}  \mathcal{L}_{\mathcal{D}}(\theta, \lambda, c)$
\STATE \textbf{Output:} $\hat \theta$, $\hat \lambda$, $\hat c$.
\end{algorithmic}
\end{algorithm}

\Cref{alg:fine_tuning_theoretical} aims to fine-tune the embedding by minimizing the loss function, which is either the binary cross entropy as defined in \eqref{eq:BCE_loss_empirical} or the squared log difference as defined in \eqref{eq:sld_loss_empirical}, which we will discuss in details in \Cref{sec:BCE} and \Cref{sec:squared_log_diff} respectively.

Note that \Cref{alg:fine_tuning_theoretical} is a theoretical version of our main algorithm. In practice, as shown in \Cref{alg:fine_tuning_practical} in \Cref{sec:exp_fine_tuning}, we will only optimize over $\theta$ by minimizing the loss while viewing $\lambda$ and $c$ as hyperparameters.

In \Cref{sec:BCE} and \Cref{sec:squared_log_diff}, we show that under both choices of loss function, the expected prediction error will converge to 0 with an $O(1/N^{1/4})$ rate, where $N$ is the size of the dataset.

\subsection{Convergence guarantee for binary cross-entropy loss}
\label{sec:BCE}
 For any $\theta \in \Theta, \lambda \in \Lambda, c \in \mathcal{C}$, we denote the (binary) cross-entropy loss function w.r.t. the prompt pair distribution $\mu$ as
 \begin{equation} 
\begin{aligned}
\label{eq:bce_loss_distributional}
    \mathcal{L}_\mu^{\text{BCE}}(\theta, \lambda, c)  = -\mathbb{E}_{(q_1, q_2)\sim \mu}[P^\star(q_1=q_2)\log P_{\theta,\lambda,c}(q_1=q_2)  + \bar P^\star(q_1=q_2)\log \bar P_{\theta,\lambda,c}(q_1=q_2) ].
\end{aligned}
\end{equation}

To recover the ground-truth parameter $(\theta^\star, \lambda^\star, c^\star)$, one only needs to solve the optimization problem 
\begin{equation}
\label{eq:minimize_bce_loss_distributional}
    \min_{\theta\in\Theta, \lambda\in\Lambda, c\in\mathcal{C}}  \mathcal{L}_\mu^{\text{BCE}}(\theta, \lambda, c).
\end{equation}
One may observe that \eqref{eq:minimize_bce_loss_distributional} is equivalent to minimizing the expected KL divergence between the ground truth probability $P^\star$ and $P_{\theta, \lambda, c}$.

Since we do not have direct access to the underlying prompt pair distribution $\mu$ or the ground-truth probability $P^\star(q_1=q_2)$, and instead are given a dataset $\mathcal{\mathcal{D}} = \{ (q_{i,1}, q_{i, 2}, p_i)\}_{i=1}^N$, where $(q_{i,1}, q_{i,2}) \stackrel{\text{i.i.d.}}{\sim} \mu$ and $p_i \in [0, 1]$ with $p_i \sim \mathcal{P}(\cdot| q_{i,1}, q_{i,2})$ and $\mathbb{E}[p_i | q_{i,1}, q_{i,2}] = P^\star(q_1=q_2)$, 
 our algorithm minimizes the empirical version of the loss function $\mathcal{L}_{\mathcal{D}}^{\text{BCE}}(\theta, \lambda, c)$, where  

\begin{equation}
\label{eq:BCE_loss_empirical}
\begin{aligned}
    \mathcal{L}_{\mathcal{D}}^{\text{BCE}}(\theta, \lambda, c) = \frac{-1}{N}\sum_{ (q_{i,1}, q_{i, 2}, p_i) \in \mathcal{D}} \left( p_i \log P_{\theta,\lambda,c}(q_{i,1}=q_{i,2})+ \bar p_i\log \bar P_{\theta,\lambda,c}(q_{i,1}=q_{i,2}) \right). 
\end{aligned}
\end{equation}
Let $(\hat \theta, \hat \lambda, \hat c)$ denote the minimizer of the empirical loss, i.e., 
\begin{align*}
    (\hat \theta, \hat \lambda, \hat c) \in \argmin_{\theta\in\Theta, \lambda\in\Lambda, c\in\mathcal{C}} \mathcal{L}_{\mathcal{D}}^{\text{BCE}}(\theta, \lambda, c), 
\end{align*}
which is exactly the output of \Cref{alg:fine_tuning_theoretical} when the loss type $L$ is BCE.

The following theorem provides a finite sample guarantee of the convergence rate of the empirical minimizer: 

\begin{theorem}[Convergence rate of \Cref{alg:fine_tuning_theoretical}, BCE loss]
\label{thm:conv_rate_BCE}
    Under \Cref{assump:realizability,assump:boundedness}, for any $\delta \in (0, 1)$, with probability at least $1 - \delta$, it holds that 
    \begin{align*}
        &\E_{(q_1, q_2) \sim \mu} \left[\left|P^\star(q_1=q_2) - P_{\hat \theta, \hat \lambda, \hat c}(q_1=q_2)\right| \right]  \leq O\left(\frac{\sqrt{L_\lambda^{-1}+B_c} \cdot (\log (|\Theta||\Lambda||\mathcal{C}|/ \delta))^{1/4}}{N^{1/4}}\right). 
    \end{align*}
\end{theorem}

\begin{remark}
    The upper bound in \Cref{thm:conv_rate_BCE} involves the logarithm of the cardinality of parameter spaces $\Theta$, $\Lambda$, and $\mathcal{C}$, which implies these sets are finite. In practice, these parameter spaces can be infinite. We assume they are finite only for ease of presentation. When these parameter spaces are infinite, we can replace the log of cardinality with the log covering number in the upper bound by the covering set argument (such as in \citet{wang_reinforcement_2020,zhu_provably_2023}).
\end{remark}

We provide the proof of \Cref{thm:conv_rate_BCE} in the subsequent subsection.

\subsubsection{Proof of \Cref{thm:conv_rate_BCE}}

Our proof strategy is similar to that of \citet{zhan2022offline,zhu2023provably}. The first step is to show that the empirical loss defined in \eqref{eq:BCE_loss_empirical} is concentrated on the actual BCE loss \eqref{eq:bce_loss_distributional} with high probability. Since the two loss functions are very close, we can approximately optimize the actual loss by optimizing the empirical version. This is formalized in \Cref{lem:concentration_BCE}.

\begin{lemma}
\label{lem:concentration_BCE}
Under \Cref{assump:realizability,assump:boundedness}, for any $\delta \in (0, 1)$, with probability at least $1 - \delta$, it holds that 
\begin{align*}
    \left| \mathcal{L}_{\mathcal{D}}^{\text{BCE}}(\theta, \lambda, c) - \mathcal{L}_{\mu}^{\text{BCE}}(\theta, \lambda, c) \right| 
     \leq  O\left((L_\lambda^{-1}+B_c)\sqrt{\frac{\log (|\Theta||\Lambda||\mathcal{C}|/ \delta)}{N}}\right) \triangleq \epsilon_{\text{stat}}^{\text{BCE}} 
\end{align*}
for any $\theta \in \Theta, \lambda \in \Lambda, c \in \mathcal{C}$.
\end{lemma}

The proof of \Cref{lem:concentration_BCE} is deferred to \Cref{sec:proof_lem_concentration_BCE}. Based on \Cref{lem:concentration_BCE}, we can further show that the empirical minimizer $(\hat \theta, \hat \lambda, \hat c)$ is also an approximate minimizer of the actual BCE loss, which is stated in \Cref{Lem:diff_in_loss_BCE}.

\begin{lemma}
\label{Lem:diff_in_loss_BCE}
Under \Cref{assump:realizability,assump:boundedness}, with probability at least $1-\delta$, it holds that 
\begin{align*}
    \mathcal{L}_{\mu}^{\text{BCE}}(\hat \theta, \hat \lambda, \hat c) -  \mathcal{L}_{\mu}^{\text{BCE}}(\theta^\star, \lambda^\star, c^\star) \leq 2  \epsilon_{\text{stat}}^{\text{BCE}}.
\end{align*}
where $\epsilon_{\text{stat}}^{\text{BCE}}$ is defined in \Cref{lem:concentration_BCE} and
\begin{align*}
    (\hat \theta, \hat \lambda, \hat c) \in \argmin_{\theta\in\Theta, \lambda\in\Lambda, c\in\mathcal{C}} \mathcal{L}_{\mathcal{D}}^{\text{BCE}}(\theta, \lambda, c). 
\end{align*}
\end{lemma}

The proof of \Cref{Lem:diff_in_loss_BCE} is deferred to \Cref{sec:proof_lem_diff_in_loss_BCE}. Given \Cref{Lem:diff_in_loss_BCE}, we can show that the induced probability by the empirical minimizer is close to the ground-truth probability in $(2,\mu)$-norm by strong convexity and thus provides an upper bound of the expected prediction error.

Finally, we provide the formal proof of \Cref{thm:conv_rate_BCE} based on \Cref{lem:concentration_BCE} and \Cref{Lem:diff_in_loss_BCE}.

\begin{proof}[Proof of \Cref{thm:conv_rate_BCE}]
    We condition on the high probability event in \Cref{Lem:diff_in_loss_BCE}. Note that by the realizability of the ground-truth probability (\Cref{assump:realizability}) and the property of binary cross-entropy, we have 
    \begin{align*}
        (\theta^\star, \lambda^\star, c^\star) \in \argmin_{\theta\in\Theta, \lambda\in\Lambda, c\in\mathcal{C}} \mathcal{L}_{\mu}^{\text{BCE}}(\theta, \lambda, c).
    \end{align*}

    For any $\theta\in\Theta, \lambda\in\Lambda, c\in\mathcal{C}$, we map $(\theta, \lambda, c)$ to a function $f_{\theta,\lambda,c}(\cdot, \cdot): \mathcal{Q} \times \mathcal{Q} \to [0, 1]$, where 
    \begin{align*}
        f_{\theta,\lambda,c}(q_1, q_2) = P_{\theta,\lambda,c}(q_1 = q_2), \ \  \forall q_1, q_2 \in \mathcal{Q}.
    \end{align*}
    Moreover, we define functional $h$ s.t. for any function $f: \mathcal{Q}\times\mathcal{Q} \to [0,1]$,  
    \begin{align*}
       h(f) =&  -\mathbb{E}_{(q_1, q_2)\sim \mu}[P^\star(q_1=q_2)\log f(q_1, q_2) \\ & \qquad \qquad \qquad + \bar P^\star(q_1=q_2)\log (1-f(q_1, q_2)) ].
    \end{align*}
    By the definition of BCE loss, $h(f_{\theta,\lambda,c}) =  \mathcal{L}_{\mu}^{\text{BCE}}(\theta, \lambda, c)$. Therefore, \Cref{Lem:diff_in_loss_BCE} translates to 
    \begin{align}
    \label{eq:loss_diff_BCE}
        h(f_{\hat \theta, \hat \lambda, \hat c}) -  h(f_{\theta^\star, \lambda^\star, c^\star}) \leq 2  \epsilon_{\text{stat}}^{\text{BCE}}.
    \end{align}
    Note that $f_{\theta^\star, \lambda^\star, c^\star}$ is still the minimizer of $h(f)$ even if $f$ does not represent an induced probability by some $\theta, \lambda, c$.
    
    We also observe that $h(f)$ is 1-strongly convex w.r.t. $f$ in $\| \cdot \|_{2, \mu}$ norm. To see why this is the case, one can calculate the second-order derivative of $h$ w.r.t. $f(q_1,q_2)$ for any $(q_1, q_2) \in \mathcal{Q} \times \mathcal{Q}$, which is
    \begin{align*}
        \frac{\partial^2 h}{\partial f^2} (q_1, q_2)  =& \frac{P^\star(q_1=q_2)}{f^2(q_1, q_2)} + \frac{\bar P^\star(q_1=q_2)}{(1-f(q_1, q_2))^2}
        \\ \geq& P^\star(q_1=q_2) + \bar P^\star(q_1=q_2) \\ =& 1,
    \end{align*}
    which demonstrates the strong convexity. Therefore, by strong convexity and the optimality of $f_{\theta^\star, \lambda^\star, c^\star}$, we can obtain that
    \begin{align*}
        h(f_{\hat \theta, \hat \lambda, \hat c}) \geq   h(f_{\theta^\star, \lambda^\star, c^\star}) + \frac{1}{2} \| f_{\theta^\star, \lambda^\star, c^\star} - f_{\hat \theta, \hat \lambda, \hat c}\|_{2, \mu}^2.
    \end{align*}
    Combining \eqref{eq:loss_diff_BCE}, we have
    \begin{align*}
        \|f_{\theta^\star, \lambda^\star, c^\star} - f_{\hat \theta, \hat \lambda, \hat c} \|_{2, \mu} \leq 2\sqrt{\epsilon_{\text{stat}}^{\text{BCE}}}.
    \end{align*}
    Finally, by Cauchy–Schwarz inequality, we can conclude
    \begin{align*}
     &\E_{(q_1, q_2) \sim \mu} \left[\left|P^\star(q_1=q_2) - P_{\hat \theta, \hat \lambda, \hat c}(q_1=q_2)\right| \right] \\ =& \|f_{\theta^\star, \lambda^\star, c^\star} - f_{\hat \theta, \hat \lambda, \hat c} \|_{1, \mu} \leq \|f_{\theta^\star, \lambda^\star, c^\star} - f_{\hat \theta, \hat \lambda, \hat c} \|_{2, \mu} \\ \leq&  2\sqrt{\epsilon_{\text{stat}}^{\text{BCE}}}.
    \end{align*}
\end{proof}

\subsection{Convergence guarantee for squared log difference loss}
\label{sec:squared_log_diff}

In this section, we analyze the convergence rate for another loss function. Define the squared log difference (SLD) loss function as follows:
\begin{equation}
\begin{aligned}
\label{eq:sld_loss_distributional}
    \mathcal{L}_\mu^{\text{SLD}}(\theta, \lambda, c)  = \mathbb{E}_{(q_1, q_2)\sim \mu}\left[\left(\log P^\star(q_1=q_2) - \log P_{\theta,\lambda,c}(q_1=q_2)\right)^2 \right].
\end{aligned}
\end{equation}

Similar to \Cref{sec:BCE}, to recover the ground-truth parameter $(\theta^\star, \lambda^\star, c^\star)$, one only needs to solve the optimization problem 
\begin{equation*}
\label{eq:minimize_sld_loss_distributional}
    \min_{\theta\in\Theta, \lambda\in\Lambda, c\in\mathcal{C}}  \mathcal{L}_\mu^{\text{SLD}}(\theta, \lambda, c).
\end{equation*}

We also define the empirical squared log difference loss as 
\begin{equation}
\label{eq:sld_loss_empirical} 
\begin{aligned} 
 \mathcal{L}_{\mathcal{D}}^{\text{SLD}}(\theta, \lambda, c) 
 = \frac{1}{N}\sum_{ (q_{i,1}, q_{i, 2}, p_i) \in \mathcal{D}} 
\left(\log p_i - \log P_{\theta,\lambda,c}(q_{i,1}=q_{i,2})\right)^2.
\end{aligned}    
\end{equation}

Since $\log 0$ is not well-defined, for theoretical analysis, we assume that each $p_i$ in the dataset $\mathcal{D}$ satisfies $p_i = P^\star(q_{i,1}=q_{i,2})$, i.e., the label is exact the ground-truth probability.

Let $(\hat \theta, \hat \lambda, \hat c)$ denote the minimizer of the empirical loss, i.e., 
\begin{align*}
    (\hat \theta, \hat \lambda, \hat c) \in \argmin_{\theta\in\Theta, \lambda\in\Lambda, c\in\mathcal{C}} \mathcal{L}_{\mathcal{D}}^{\text{SLD}}(\theta, \lambda, c), 
\end{align*}
which is exactly the output of \Cref{alg:fine_tuning_theoretical} when the loss type $L$ is SLD.

Now, we provide a finite sample convergence guarantee for the expected prediction error under the squared log difference loss: 

\begin{theorem}[Convergence rate of \Cref{alg:fine_tuning_theoretical}, SLD loss]
\label{thm:conv_rate_sld}
    Under \Cref{assump:realizability,assump:boundedness}, for any $\delta \in (0, 1)$, with probability at least $1 - \delta$, it holds that 
    \begin{align*}
        \E_{(q_1, q_2) \sim \mu} \left[\left|P^\star(q_1=q_2) - P_{\hat \theta, \hat \lambda, \hat c}(q_1=q_2)\right| \right] 
        \leq O\left(\frac{(L_\lambda^{-1}+B_c) \cdot (\log (|\Theta||\Lambda||\mathcal{C}|/ \delta))^{1/4}}{N^{1/4}}\right). 
    \end{align*}
\end{theorem}

The proof of \Cref{thm:conv_rate_sld} is similar to the proof of \Cref{thm:conv_rate_BCE} and is deferred to \Cref{sec:proof_sld}.

\begin{remark}
    Compared to the bound for BCE loss, there is an additional $\sqrt{L_\lambda^{-1} + B_c}$  factor in the bound for SLD loss. This implies that BCE loss is more sample-efficient than SLD loss when the lower bound of $\lambda \in \Lambda$, $L_\lambda$, is small or the upper bound of $c \in \mathcal{C}$, $B_c$, is large. This theoretical result is consistent with our empirical results in \Cref{tab:efficiency}, where the BCE model achieves better caching efficiency than the SLD model. More details will be discussed in \Cref{sec:exp_caching_sim}.
\end{remark}

%% file: exp.tex
\newcommand{\numBadResponse}{\texttt{nBadResponse}}
\newcommand{\numLLMQuery}{\texttt{nLLMQuery}}
\newcommand{\throupghput}{\texttt{thp}}
\newcommand{\numCorrectHit}{\texttt{nCorrectHit}}
\newcommand{\numFalseHit}{\texttt{nFalseHit}}
\newcommand{\efficiency}{\texttt{efficiency}}
\newcommand{\numExpHit}{\texttt{nExpectedHit}}

\section{Experiments}
\label{sec:exp}

In this section, we show experimental results that our distillation-based fine-tuning method can improve the accuracy of caching prediction and improve the caching efficiency for prompt streaming with caching. 

We first introduce how we construct the training dataset in \Cref{sec:exp_construct_dataset}. The dataset we constructed is a hard dataset by careful design, and the previous embedding model in \citet{wang2022text} only achieves a 0.51 AUC on our dataset.

We then in \Cref{sec:exp_fine_tuning} fine-tune the embedding from \citet{wang2022text} using \Cref{alg:fine_tuning_practical}, which is the practical version of \Cref{alg:fine_tuning_theoretical} we analyzed in \Cref{sec:theory}. The resulting models under both BCE loss and SLD loss achieve a 0.81 AUC on the validation set, which significantly improves over the initial model. 

Furthermore, in \Cref{sec:exp_caching_sim}, we conduct a simulation on prompt streaming with a cache and compare different embedding models as the predictor of caching hit. As \Cref{tab:efficiency} shows, our BCE model and SLD model achieve caching efficiency of $54.0\%$ and $52.4\%$, respectively, which both improve over the efficiency of $46.0\%$ of the previous model. 

\subsection{Construction of the dataset}
\label{sec:exp_construct_dataset}

We first extract all prompts (queries) from the \texttt{natural\_questions} dataset~\citep{47761} and discard the responses. For each prompt, we compute a vector embedding using the last layer of the \texttt{intfloat/e5-large-v2} model~\citep{wang2022text}. We also deleted repeated prompts. 

Now for each prompt, we search the three nearest neighbors using FAISS~\citep{johnson2019billion}, where the distance of a prompt pair is defined by the cosine similarity of the two corresponding embedding vectors as in \Cref{defn:induced_prob}. We then sample 70k prompts uniformly at random, and for each prompt, we obtain three prompt pairs forming by the prompt itself and each of their three nearest neighbors. 
Therefore, we get 210k prompt pairs in total, and we use GPT-4~\citep{openai2023gpt4} to label whether each prompt pair can be answered by the same response (0 or 1, where 1 means they can be answered by the same response). The exact prompt we use to label each pair is presented in \Cref{sec:app_prompt_to_label}.

\begin{algorithm}[h]
\caption{Constructing a hard dataset}
\label{alg:construct_dataset}
\begin{algorithmic}[1]
\STATE \textbf{Input:} Initial large dataset $\mathcal{D}_0 = \{ (q_{i,1}, q_{i, 2}, p_i)\}_{i=1}^{N_0}$ where $p_i = 0$ or $1$ is the label, an existing embedding model $v_\theta(\cdot)$ (in our case, $v_\theta(\cdot)$ is \texttt{intfloat/e5-large-v2}).
\FOR{$n = 1, 2, \ldots, N_0$}
\STATE $c_i \gets \texttt{sim} ( v_\theta(q_{i,1}), v_\theta(q_{i,2}))$  
\ENDFOR

\STATE{Sort data points in $\mathcal{D}_0$ by the ascending order of $c_i$}

\STATE{$\mathcal{D} \gets \{\}$}
\STATE {LastLabel $\gets 1$}
\FOR{$n = 1, 2, \ldots, N_0$}
\IF{$p_i \neq $ LastLabel}
\STATE $\mathcal{D} \gets \mathcal{D} \cup \{ (q_{i,1}, q_{i, 2}, p_i) \}$
\STATE LastLabel $\gets p_i$
\ENDIF
\ENDFOR

\STATE \textbf{Output:}  the hard dataset $\mathcal{D}$ 
\end{algorithmic}
\end{algorithm}

The next step is to select prompt pairs among the above 210k labeled pairs to obtain the final dataset, which is not too small in size, and the existing embedding model \texttt{intfloat/e5-large-v2} has a poor AUC (e.g., close to 0.5) on the selected dataset. Note that the existence of such a dataset implies that \texttt{intfloat/e5-large-v2} is not an ideal embedding model for predicting whether two prompts can be answered by the same response.~\footnote{If an existing embedding model $v_\theta(\cdot)$ is nearly perfect, it would be hard to find two prompt pairs $(q_{11}, q_{12})$ with label 1, and $(q_{21}, q_{22})$ with label 0, s.t. $\texttt{sim}(v_\theta(q_{11}), v_\theta(q_{12})) < \texttt{sim}(v_\theta(q_{21}), v_\theta(q_{22}))$. In that case, even if we can find some ``bad examples'' to construct a hard dataset, the size of the dataset will be very small.} \Cref{alg:construct_dataset} presents how we construct such a dataset.

In \Cref{alg:construct_dataset}, we first sort all data points by cosine similarity. Then, we add data points in ascending order of similarity and ensure the added prompt pairs have alternate labels. By the construction procedure, the AUC of \texttt{intfloat/e5-large-v2} on the constructed hard dataset $\mathcal{D}$ is approximately 0.5. In our experiments, the size of $\mathcal{D}$ is 37382, which is large enough for our training and implies that the existing embedding \texttt{intfloat/e5-large-v2} might not be an ideal model for caching prediction.

Finally, we show two examples of prompt pairs in our dataset, where the first one has label 0 but high cosine similarity according to \texttt{intfloat/e5-large-v2}, and the second one has label 1 but low cosine similarity.

\begin{mdframed}[backgroundcolor=gray!20]
Example 1:

Prompt 1: when did sat change from 2400 to 1600

Prompt 2: when did sat change from 1600 to 2400

Label: 0

Cosine similarity: 0.9973602294921875
\end{mdframed}

\begin{mdframed}[backgroundcolor=gray!20]
Example 2:

Prompt 1: who wrote a song for you leon russell or donny hathaway

Prompt 2: who wrote the song a song for you

Label: 1

Cosine similarity: 0.8098968863487244
\end{mdframed}

\subsection{Fine-tuning embeddings}
\label{sec:exp_fine_tuning}

We split the dataset constructed in \Cref{sec:exp_construct_dataset} into training, validation, and test sets with a ratio of 7:1:2.

We fine-tune using the above training set from embedding \texttt{intfloat/e5-large-v2} under BCE loss or SLD loss. Slightly different from the theoretical version \Cref{alg:fine_tuning_theoretical}, in the practical version \Cref{alg:fine_tuning_practical}, we view $\lambda$ and $c$ as hyperparameters, set $\lambda = 0.01$ for both loss functions, and set $c = 88$ for BCE loss and $c = 90$ for SLD loss. For SLD loss, we clip the label to $[10^{-10}, 1]$ to avoid calculating $\log 0$, which is not well-defined. We use AdamW~\citep{loshchilov2017decoupled} as our optimizer and set the learning rate $\eta$ to be $10^{-5}$. The full list of hyperparameter values used in our experiments is shown in \Cref{sec:app_hyperparameter}.

\begin{algorithm}[h]
\caption{Supervised embedding fine-tuning (practical version)}
\label{alg:fine_tuning_practical}
\begin{algorithmic}[1]
\STATE \textbf{Input:} Training dataset $\mathcal{\mathcal{D}} = \{ (q_{i,1}, q_{i, 2}, p_i)\}_{i=1}^N$, hyperparameters $\lambda$, $c$, embedding model $v_\theta(\cdot)$, learning rate $\eta$, number of epochs $E$, batch size $B$, loss type $L$
\STATE Initialization: $\theta \gets \theta_0$.
\IF{L = SLD}
\FOR{$i = 1, 2, \ldots, N$ }
\STATE $p_i \gets \texttt{clip}(p_i, 10^{-10}, 1)$ 
\ENDFOR
\ENDIF
\FOR{$e = 1, 2, \ldots, E$}
\FOR{$b = 1, 2, \ldots, N/B$}
\STATE Let $\mathcal{B}$ be a mini-batch sampled from $\mathcal{D}$ of size B
\IF{$L$ = BCE}
\STATE  $\mathcal{L}_{\mathcal{B}} \gets \mathcal{L}^{\text{BCE}}_{\mathcal{B}}$ as defined in \eqref{eq:BCE_loss_empirical}
\ELSIF{$L$ = SLD}
\STATE $\mathcal{L}_{\mathcal{B}} \gets \mathcal{L}^{\text{SLD}}_{\mathcal{B}}$ as defined in \eqref{eq:sld_loss_empirical}
\ELSE
\STATE Raise Not Implemented Error
\ENDIF
\STATE Perform single step to update $\theta$ by AdamW using loss $\mathcal{L}_{\mathcal{B}}$ and learning rate $\eta$
\ENDFOR
\ENDFOR
\STATE \textbf{Output:} $v_\theta(\cdot)$
\end{algorithmic}
\end{algorithm}

\begin{figure*}[htbp]
    \centering
    \subfloat[ROC before fine-tuning (AUC = 0.51)]{
        \includegraphics[width=0.3\textwidth]{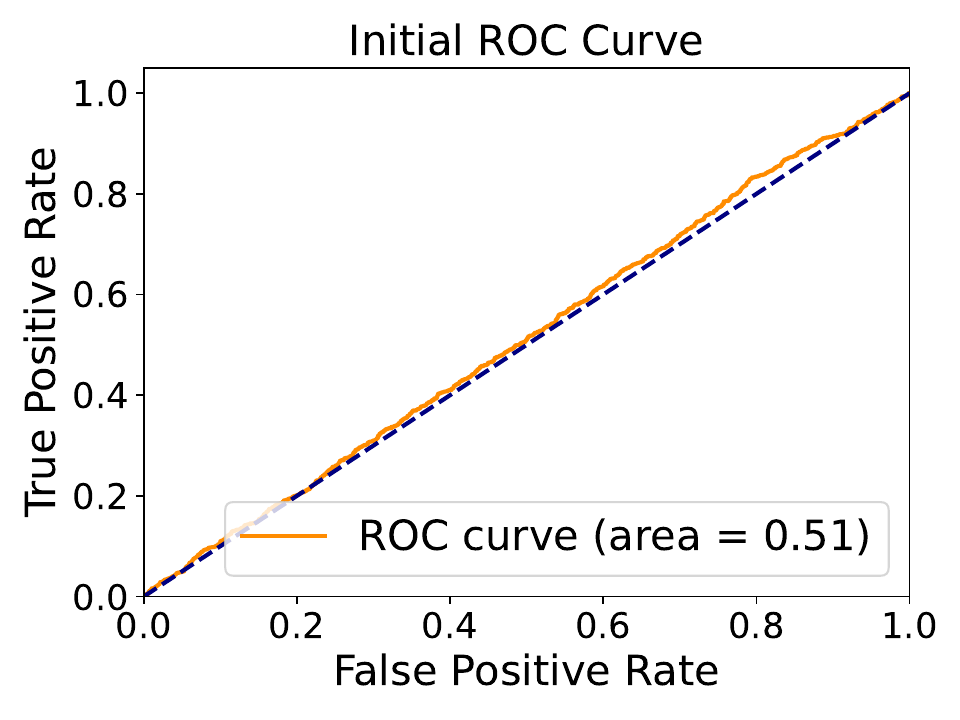}
        \label{fig:roc_epoch0}
    }
    \hfill
    \subfloat[ROC after Epoch 12 using BCE loss (AUC = 0.81)]{
        \includegraphics[width=0.3\textwidth]{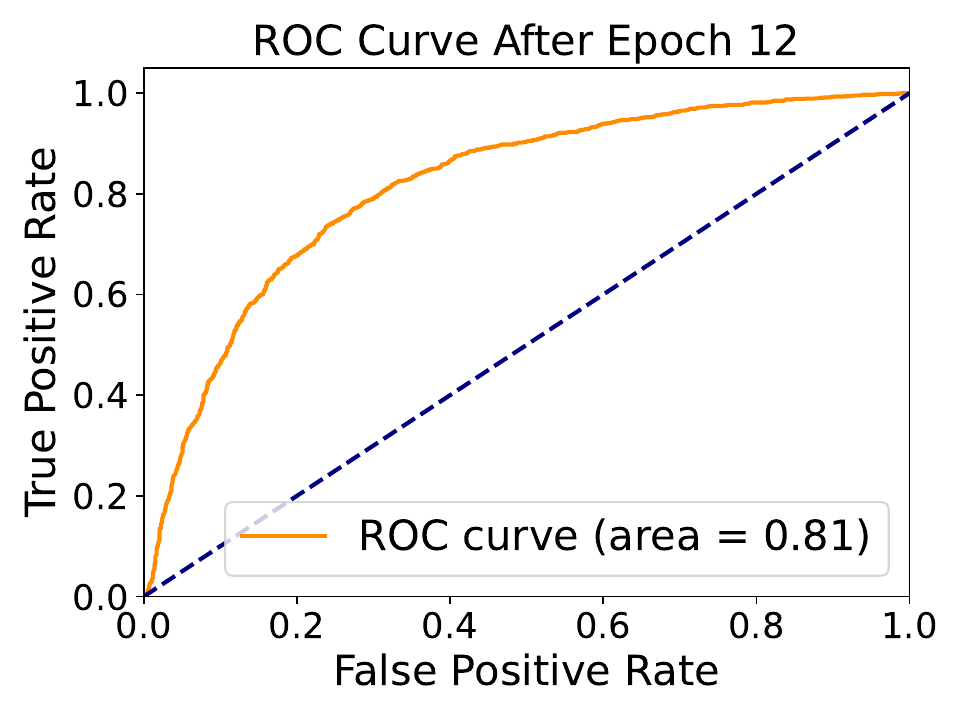}
        \label{fig:roc_epoch8_cross}
    }
    \hfill
    \subfloat[ROC after Epoch 6 using SLD loss (AUC = 0.81)]{
        \includegraphics[width=0.3\textwidth]{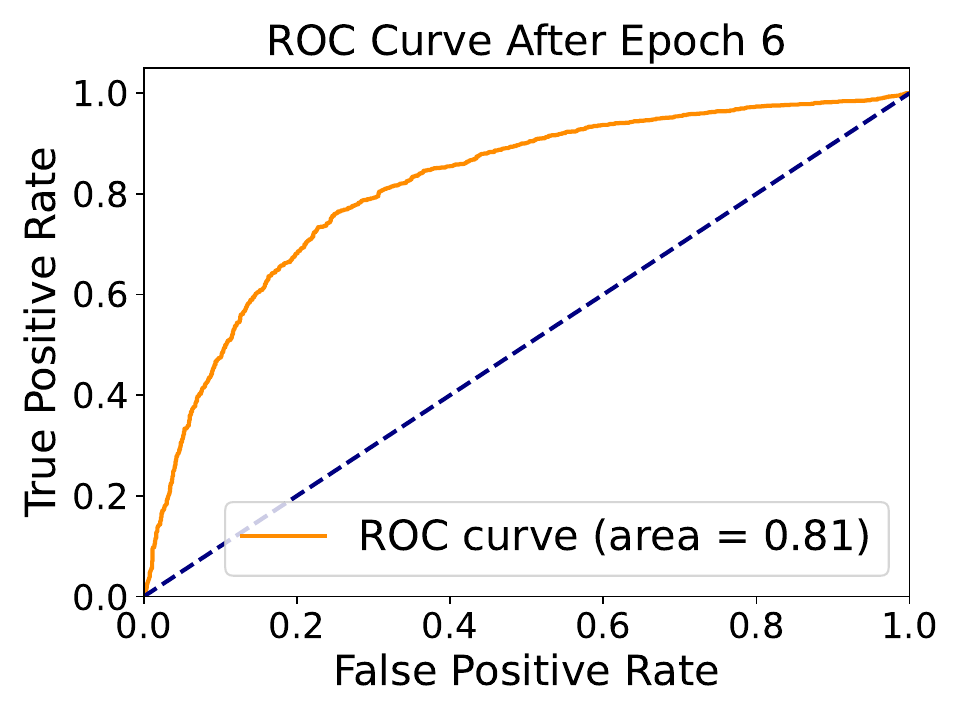}
        \label{fig:roc_epoch8_squared}
    }
    \caption{Comparison of ROC curves.}
    \label{fig:roc_curves}
\end{figure*}

We plot the ROC curve as well as AUC on the validation set for the initial model (i.e., \texttt{intfloat/e5-large-v2}), BCE model and SLD model trained by \Cref{alg:fine_tuning_practical} in \Cref{fig:roc_curves}, which shows that fine-tuning on our constructed dataset using either loss function can help to improve the AUC.

\subsection{Simulation of the prompt streaming with caching}
\label{sec:exp_caching_sim}

We also conduct a simulation to validate that the caching through embedding similarity with our fine-tuned embedding model can improve over the initial embedding model (i.e., the \texttt{intfloat/e5-large-v2} model in our case). We first create the prompt streaming dataset, which contains 1000 prompts. To be more specific, we randomly sample 250 prompt pairs with label 0 in the test set, and also randomly sample 250 prompt pairs with label 1. Therefore, we have 500 prompt pairs, and we decompose them into 1000 prompts in random order. Since the main focus of this paper is not to study the tradeoff between the size of the cache and the caching efficiency, we assume for simplicity that the cache has unlimited size. 

\begin{table*}[htbp]
    \centering
        \renewcommand{\arraystretch}{1.1} %
    \begin{tabular}{ccccccccc}
        \toprule
         threshold $\tau$ & 0.88 & 0.89 & 0.90 & 0.91 & 0.92 & 0.93 & 0.94  \\
        \midrule
        Initial model efficiency &  38.0\% & \textbf{46.0\%} & 41.6\% & 40.0\% & \textbf{46.0\%} & 38.0\% & 35.2\% \\
        \bottomrule
        \\
    \end{tabular}
    \begin{tabular}{ccccccccc}
        \toprule
         threshold $\tau$ & 0.88 & 0.89 & 0.90 & 0.91 & 0.92 & 0.93 & 0.94  \\
        \midrule
        BCE model efficiency & 23.6\% & 36.4\% & 45.6\% & 50.8\% & \textbf{54.0\%} & 53.2\% & 44.4\% \\
        \bottomrule
        \\
    \end{tabular}
    \begin{tabular}{ccccccccc}
        \toprule
         threshold $\tau$ & 0.78 & 0.80 & 0.82 & 0.84 & 0.86 & 0.88 & 0.90  \\
        \midrule
    
        SLD model efficiency  & 32.8\% & 37.2\% & \textbf{52.4\%} & 50.4\% & 47.2\% & 44.0\% & 36.4\% \\
        \bottomrule
        \\
    \end{tabular}
    \caption{Comparison of the caching efficiency of different embedding models. The efficiency is defined as ($\#$ correct caching hit - $\#$ false caching hit)/($\#$ of expected caching hit). In our test, we have 1000 prompts in the streaming, which come from 500 prompt pairs, of which 250 pairs with label one. So $\#$ of expected caching hit here is 250.}
    \label{tab:efficiency}
\end{table*}

\begin{algorithm}[h]
\caption{Simulation of prompt streaming with cache (of unlimited size)}
\label{alg:caching_sim}
\begin{algorithmic}[1]
\STATE \textbf{Input:} Prompt streaming $\{q_t \}_{t=1}^T$, embedding model $v(\cdot)$, threshold $\tau$, number of expected caching hit $\texttt{nE}$, cache $\mathcal{C}$, language model $\texttt{LM}(\cdot)$
\STATE Initialization: $\mathcal{C} \gets \{\}, \texttt{nC} \gets 0, \texttt{nF} \gets 0$.
\FOR{$t = 1, 2, \ldots, T$}
\STATE Receive the prompt $q_t$
\STATE Find the nearest neighbor $(q, v(q), r)$ of $q_t$ in $\mathcal{C}$, where the distance is defined as $1 - \texttt{sim}(v(q_t), v(q))$
\IF{$\mathcal{C}$ is empty or $\texttt{sim}(v(q), v(q_t)) \leq \tau$}
\STATE $r \gets \texttt{LM}(q_t),\ \ $ $ \mathcal{C} \gets \mathcal{C} \cup \{ (q_t, v(q_t), r) \}$
\ELSE
\IF{$r$ is a good response of $q_t$}
\STATE $\texttt{nC} \gets \texttt{nC} + 1$
\ELSE 
\STATE $\texttt{nF} \gets \texttt{nF} + 1$
\ENDIF
\ENDIF
\ENDFOR
\STATE \textbf{Output:} $(\texttt{nC} - \texttt{nF})/\texttt{nE}$
\end{algorithmic}
\end{algorithm}

The cache is initialized to be empty. We maintain two counters: $\numCorrectHit$ and $\numFalseHit$, which represent the number of correct caching hits and the number of false caching hits, respectively.  At each time, a prompt $q$ from the streaming dataset arrives, and its embedding $v(q) \in \mathbb{R}^d$ (in our experiments, $d = 1024$) is calculated. We first find the nearest neighbor in the cache, which is a tuple $(q_0, v(q_0), r_0)$, where $q_0$ is a prompt, $r_0$ is the response, and $v(q_0)$ is the embedding. The distance is measured by cosine similarity. If  $\texttt{sim} (v(q_1), v(q_2)) > \tau$ where $\tau \in [0,1]$ is a threshold, we view it as a caching hit and use $r_0$ as the response of $q$; otherwise, we view it as a caching miss, directly query the LLM to get the response $r$, and add the tuple $(q, v(q), r)$ to the cache. 

In order to test the caching efficiency, whenever there is a caching hit, we will query GPT4 whether $r_0$ is a good response to $q$. (The exact prompt we used to query GPT4 is in \Cref{sec:app_prompt_to_check}.) If it is a correct caching hit, we set $\numCorrectHit \gets \numCorrectHit + 1$; if not, we set $\numFalseHit \gets \numFalseHit + 1$. 

Finally, we calculate the caching efficiency as
\begin{align*}
    \efficiency = \frac{\numCorrectHit - \numFalseHit}{\numExpHit},
\end{align*}
where $\numExpHit = 250$ in our experiments as discussed at the beginning of this subsection. The whole simulation procedure is formalized in \Cref{alg:caching_sim}.

We choose $v$ to be one of the three models:  the initial model \texttt{intfloat/e5-large-v2}, BCE model (fine-tuned for 12 epochs), and SLD model (fine-tuned for 6 epochs) as in \Cref{sec:exp_fine_tuning}. We also choose different threshold $\tau$ and pick the best one for each embedding model. The result is presented in \Cref{tab:efficiency}, which shows that after fine-tuning using either loss, the caching efficiency is improved over the initial model without fine-tuning.

%% file: conclusion.tex
\section{Conclusions}
\label{sec:conc}

In this paper, we study efficient prompt caching for LLMs by modeling the ground-truth probability of whether a prompt pair can be answered by the same response via embedding similarity, and fine-tuning existing semantic embeddings on our newly constructed dataset. We provide both theoretical guarantee and empirical evidence that our proposed distillation-based method can improve the accuracy of caching prediction and efficiency. Interesting future directions include improving the $O(1/N^{1/4})$ rate and proposing novel training objectives that might result in better embedding models.

%% file: appendix.tex
\section{Missing Proofs}
\label{sec:proof}

\subsection{Proof of \Cref{lem:concentration_BCE}}
\label{sec:proof_lem_concentration_BCE}

\begin{proof}[Proof of \Cref{lem:concentration_BCE}]
    We first consider any fixed $\theta \in \Theta, \lambda \in \Lambda, c \in \mathcal{C}$. One can observe that $\mathcal{L}_{\mathcal{D}}^{\text{BCE}}(\theta, \lambda, c)$ is an unbiased estimator of $\mathcal{L}_{\mu}^{\text{BCE}}(\theta, \lambda, c)$ since
    \begin{align*}
        & \mathbb{E}[\mathcal{L}_{\mathcal{D}}^{\text{BCE}}(\theta, \lambda, c)] \\ 
        =& - \mathbb{E}_{(q_1,q_2)\sim \mu, p \sim \mathcal{P}(\cdot| q_1, q_2)} \left[ p \log P_{\theta,\lambda,c}(q_{1}=q_{2}) + \bar p\log \bar P_{\theta,\lambda,c}(q_{1}=q_{2}) \right] \\
        =& - \mathbb{E}_{(q_1,q_2)\sim \mu} \left[ \mathbb{E}_{p \sim \mathcal{P}(\cdot| q_1, q_2)} \left[ p \log P_{\theta,\lambda,c}(q_{1}=q_{2}) + \bar p\log \bar P_{\theta,\lambda,c}(q_{1}=q_{2})  | q_1, q_2 \right] \right]
        \\ =& - \mathbb{E}_{(q_1,q_2)\sim \mu}\left[ P^\star(q_1=q_2) \log P_{\theta,\lambda,c}(q_{1}=q_{2}) + \bar P^\star(q_1=q_2)\log \bar P_{\theta,\lambda,c}(q_{1}=q_{2})  \right]
        \\ =& \mathcal{L}_\mu^{\text{BCE}}(\theta, \lambda, c),
    \end{align*}
    where the first equality holds since the data points in the dataset are i.i.d. distributed, the second equality holds due to tower property, and the third equality holds by the linearity of expectation.

    Also, we note that the empirical loss for each data point can be upper bounded by 
    \begin{align*}
    &\left| p_i \log P_{\theta,\lambda,c}(q_{i,1}=q_{i,2}) + \bar p_i\log \bar P_{\theta,\lambda,c}(q_{i,1}=q_{i,2}) \right| \\ 
    \leq& \max\left\{ \left| \log P_{\theta,\lambda,c}(q_{i,1}=q_{i,2}) \right| , \left| \log \bar P_{\theta,\lambda,c}(q_{i,1}=q_{i,2}) \right|  \right\} \\ 
    =&  \log \left( \max\left\{ \frac{1}{P_{\theta,\lambda,c}(q_{i,1}=q_{i,2})} , \frac{1}{1 - P_{\theta,\lambda,c}(q_{i,1}=q_{i,2})} \right\}\right) .
    \end{align*}
    Since $\sigmoid(-x) = 1 - \sigmoid(x)$
    and $\texttt{sim} ( v_\theta(q_1), v_\theta(q_2)) / \lambda - c \in [-L_\lambda^{-1}-B_c, L_\lambda^{-1}+B_c]$ by \Cref{assump:boundedness}, we can obtain that 
    \begin{align*}
        \frac{1}{P_{\theta,\lambda,c}(q_1 = q_2)} = \frac{1}{\sigmoid(\texttt{sim} ( v_\theta(q_1), v_\theta(q_2)) / \lambda - c)} \leq  1+ \exp(L_\lambda^{-1}+B_c).
    \end{align*}
    By the symmetry of $\sigmoid(\cdot)$ and the range of $\texttt{sim} ( v_\theta(q_1), v_\theta(q_2)) / \lambda - c$, it also holds that 
    \begin{align*}
        \frac{1}{1 - P_{\theta,\lambda,c}(q_1 = q_2)} \leq 1 + \exp(L_\lambda^{-1}+B_c).
    \end{align*}
    Therefore, 
    \begin{align*}
        & \left| p_i \log P_{\theta,\lambda,c}(q_{i,1}=q_{i,2}) + \bar p_i\log \bar P_{\theta,\lambda,c}(q_{i,1}=q_{i,2}) \right| \\ \leq& \log\left(1 + \exp(L_\lambda^{-1}+B_c)\right) \leq O(L_\lambda^{-1}+B_c).
    \end{align*} 
    By Hoeffding's inequality, we have with probability at least $1-\delta$, it holds that 
    \begin{align*}
        \left| \mathcal{L}_{\mathcal{D}}^{\text{BCE}}(\theta, \lambda, c) - \mathcal{L}_{\mu}^{\text{BCE}}(\theta, \lambda, c) \right| \leq O\left((L_\lambda^{-1}+B_c)\sqrt{\frac{\log (1/ \delta)}{N}}\right).
    \end{align*}
    Applying a union bound over all $\theta \in \Theta, \lambda \in \Lambda, c \in \mathcal{C}$ concludes the result.
\end{proof}

\subsection{Proof of \Cref{Lem:diff_in_loss_BCE}}
\label{sec:proof_lem_diff_in_loss_BCE}

\begin{proof}[Proof of \Cref{Lem:diff_in_loss_BCE}]
       We condition on the high probability event in \Cref{lem:concentration_BCE}. Note that 
    \begin{align*}
        &\mathcal{L}_{\mu}^{\text{BCE}}(\hat \theta, \hat \lambda, \hat c) -  \mathcal{L}_{\mu}^{\text{BCE}}(\theta^\star, \lambda^\star, c^\star) \\ =& \underbrace{\mathcal{L}_{\mu}^{\text{BCE}}(\hat \theta, \hat \lambda, \hat c)  - \mathcal{L}_{\mathcal{D}}^{\text{BCE}}(\hat \theta, \hat \lambda, \hat c) }_{(1)} + \underbrace{\mathcal{L}_{\mathcal{D}}^{\text{BCE}}(\hat \theta, \hat \lambda, \hat c) - \mathcal{L}_{\mathcal{D}}^{\text{BCE}}(\theta^\star, \lambda^\star, c^\star)}_{(2)} \\ &+  \underbrace{\mathcal{L}_{\mathcal{D}}^{\text{BCE}}(\theta^\star, \lambda^\star, c^\star) - \mathcal{L}_{\mu}^{\text{BCE}}(\theta^\star, \lambda^\star, c^\star)}_{(3)}.
    \end{align*}
    $(1), (3) \leq \epsilon_{\text{stat}}^{\text{BCE}}$ by \Cref{lem:concentration_BCE} and $(2) \leq 0$ by the optimality of $(\hat \theta, \hat \lambda, \hat c)$, which completes the proof.
\end{proof}

\subsection{Proof of \Cref{thm:conv_rate_sld}}
\label{sec:proof_sld}

\begin{proof}[Proof of \Cref{thm:conv_rate_sld}]
    The proof is similar to the proof of \Cref{thm:conv_rate_BCE}. First, it is easy to see that the empirical loss  $\mathcal{L}_{\mathcal{D}}^{\text{SLD}}(\theta, \lambda, c)$ is an unbiased estimator of $\mathcal{L}_\mu^{\text{SLD}}(\theta, \lambda, c)$ by definition since $p_i =  P^\star(q_{i,1}=q_{i,2})$. Also, 
    \begin{align*}
        \left(\log p_i - \log P_{\theta,\lambda,c}(q_{i,1}=q_{i,2})\right)^2 \leq \left(\log\left(1 + \exp(L_\lambda^{-1}+B_c)\right)\right)^2 = O\left( (L_\lambda^{-1}+B_c)^2\right).
    \end{align*}
    Therefore, by Hoeffding's inequality and union bound, we have that with probability at least $1-\delta$, it holds that 
    \begin{align*}
        \left| \mathcal{L}_{\mathcal{D}}^{\text{SLD}}(\theta, \lambda, c) - \mathcal{L}_{\mu}^{\text{SLD}}(\theta, \lambda, c) \right| \leq O\left((L_\lambda^{-1}+B_c)^2\sqrt{\frac{\log (|\Theta||\Lambda||\mathcal{C}|/ \delta)}{N}}\right) \triangleq \epsilon_{\text{stat}}^{\text{SLD}}.
    \end{align*}
    Similar to \Cref{Lem:diff_in_loss_BCE}, we can obtain that 
    \begin{align}
    \label{eq:loss_diff_sld}
    \mathcal{L}_{\mu}^{\text{SLD}}(\hat \theta, \hat \lambda, \hat c) -  \mathcal{L}_{\mu}^{\text{SLD}}(\theta^\star, \lambda^\star, c^\star) \leq 2  \epsilon_{\text{stat}}^{\text{SLD}}.
    \end{align}
    Now, for any $\theta\in\Theta, \lambda\in\Lambda, c\in\mathcal{C}$, we map $(\theta, \lambda, c)$ to a function $f_{\theta,\lambda,c}(\cdot, \cdot): \mathcal{Q} \times \mathcal{Q} \to [0, +\infty)$, where 
    \begin{align*}
        f_{\theta,\lambda,c}(q_1, q_2) = - \log P_{\theta,\lambda,c}(q_1 = q_2), \ \  \forall q_1, q_2 \in \mathcal{Q}.
    \end{align*}
    Moreover, we define functional $h$ s.t. for any function $f: \mathcal{Q}\times\mathcal{Q} \to [0,1]$,  
    \begin{align*}
       h(f) =&  \mathbb{E}_{(q_1, q_2)\sim \mu}[(f(q_1, q_2) + \log P^\star(q_1 = q_2))^2].
    \end{align*}
    By the definition of squared log difference loss, $h(f_{\theta,\lambda,c}) =  \mathcal{L}_{\mu}^{\text{SLD}}(\theta, \lambda, c)$. Therefore, \eqref{eq:loss_diff_sld} translates to 
    \begin{align}
    \label{eq:loss_diff_sld_functional}
        h(f_{\hat \theta, \hat \lambda, \hat c}) -  h(f_{\theta^\star, \lambda^\star, c^\star}) \leq 2  \epsilon_{\text{stat}}^{\text{SLD}}.
    \end{align}
    Note that $f_{\theta^\star, \lambda^\star, c^\star}$ is still the minimizer of $h(f)$ even if $f$ cannot be induced by some $\theta, \lambda, c$.
    
    We also observe that $h(f)$ is 2-strongly convex w.r.t. $f$ in $\| \cdot \|_{2, \mu}$ norm by calculating the second-order derivative of $h$ w.r.t. $f$. Therefore, by strong convexity and the optimality of $f_{\theta^\star, \lambda^\star, c^\star}$, we can obtain that
    \begin{align*}
        h(f_{\hat \theta, \hat \lambda, \hat c}) \geq   h(f_{\theta^\star, \lambda^\star, c^\star}) + \| f_{\theta^\star, \lambda^\star, c^\star} - f_{\hat \theta, \hat \lambda, \hat c}\|_{2, \mu}^2.
    \end{align*}
    Combining \eqref{eq:loss_diff_sld_functional}, we can obtain 
    \begin{align*}
        \mathbb{E}_{(q_1, q_2)\sim \mu}\left[\left(\log P^\star(q_1=q_2) - \log P_{\theta,\lambda,c}(q_1=q_2)\right)^2 \right] = \| f_{\theta^\star, \lambda^\star, c^\star} - f_{\hat \theta, \hat \lambda, \hat c}\|_{2, \mu}^2 \leq 2  \epsilon_{\text{stat}}^{\text{SLD}}.
    \end{align*}
    Note that by mean value theorem, for any $0 < x < y < 1$, $\log x - \log y = (x-y)/z$ for some $z \in (x, y)$. Therefore, $(\log x - \log y)^2 = (x-y)^2/z^2 > (x-y)^2$. This implies
    \begin{align*}
       &\mathbb{E}_{(q_1, q_2)\sim \mu}\left[\left(P^\star(q_1=q_2) -  P_{\theta,\lambda,c}(q_1=q_2)\right)^2 \right] \\ \leq& \mathbb{E}_{(q_1, q_2)\sim \mu}\left[\left(\log P^\star(q_1=q_2) - \log P_{\theta,\lambda,c}(q_1=q_2)\right)^2 \right] = 2  \epsilon_{\text{stat}}^{\text{SLD}}.
    \end{align*}
    By Cauchy–Schwarz inequality, we can conclude
    \begin{align*}
     &\E_{(q_1, q_2) \sim \mu} \left[\left|P^\star(q_1=q_2) - P_{\hat \theta, \hat \lambda, \hat c}(q_1=q_2)\right| \right] \\ \leq& \sqrt{\mathbb{E}_{(q_1, q_2)\sim \mu}\left[\left(P^\star(q_1=q_2) -  P_{\theta,\lambda,c}(q_1=q_2)\right)^2 \right]}  \leq  \sqrt{2\epsilon_{\text{stat}}^{\text{SLD}}}.
    \end{align*}
\end{proof}

\section{Details of Prompts Used in Experiments}
\label{sec:app_prompt_details}

\subsection{Prompt used to label prompt pairs}
\label{sec:app_prompt_to_label}

Below we show the exact prompt that we used to label prompt pairs. ``PROMPT1'' and ``PROMPT2'' are placeholders which will be replaced by the prompt pair we want to label.

\begin{singleframed}[backgroundcolor=gray!20]
\setlength{\parindent}{\defparindent}
In the following three tasks, you are given two prompts for each task. You need to tell whether the two prompts can be responded by the same response. Please only answer Yes or No without any explanation.
\\
\\
Task 1:
\\
Prompt 1: does minho die in the scorch trials book
\\
Prompt 2: does thomas die in the death cure book
\\
Answer: No
\\
\\
Task 2:
\\
Prompt 1: how many passengers on titanic when it sank
\\
Prompt 2: How many passengers were aboard the Titanic when it went down
\\
Answer: Yes
\\
\\
Task 3:
\\
Prompt 1: PROMPT1
\\
Prompt 2: PROMPT2
\\
Answer:
\end{singleframed}

\subsection{Prompt used to test whether a given response is valid for a given prompt}
\label{sec:app_prompt_to_check}

Below we show the exact prompt that we used to check whether a given response is valid for a given prompt. ``PLACEHOLDER\_PROMPT'' and ``PLACEHOLDER\_RESPONSE''  are placeholders that will be replaced by the actual prompt and response that we aim to check.

\begin{singleframed}[backgroundcolor=gray!20]
\setlength{\parindent}{\defparindent}
In the following task, you will be given a prompt and a response. You are required to evaluate whether the response is a good response for the given prompt. The prompt starts after 'PROMPT:', the response starts after 'RESPONSE:', and your answer starts after 'ANSWER:'. Your answer should start with the explanation, and finally output [[Yes]] if it is a good response, or [[No]] if it is not. Now the prompt begins.
\\
\\
PROMPT: PLACEHOLDER\_PROMPT
\\
\\
RESPONSE: PLACEHOLDER\_RESPONSE
\\
\\
ANSWER:
\end{singleframed}

\section{Hyperparameters for \Cref{alg:fine_tuning_practical}}
\label{sec:app_hyperparameter}

\Cref{tab:hyperparameters} lists the value of hyperparameters we used in our main algorithm \Cref{alg:fine_tuning_practical}.

\begin{table*}[htbp]
    \centering
        \renewcommand{\arraystretch}{1.1} %
    \begin{tabular}{cc}
        \toprule
         Hyperparameters & Value  \\
        \midrule
        learning rate $\eta$ & $10^{-5}$ \\
        \midrule
        number of epochs $E$ & $20$ \\
        \midrule
         $\lambda$ & $0.01$ \\
         \midrule
         $c$ (BCE loss) & 88 \\
        \midrule
        $c$ (SLD loss) & 90 \\
        \midrule
        batch size $B$ & 16 \\
        \midrule
        AdamW hyperparameters & default \\
        \bottomrule
        \\
    \end{tabular}
    \caption{Value of hyperparameters for \Cref{alg:fine_tuning_practical}}
    \label{tab:hyperparameters}
\end{table*}

%% file: main.bbl
\begin{thebibliography}{45}
\providecommand{\natexlab}[1]{#1}
\providecommand{\url}[1]{\texttt{#1}}
\expandafter\ifx\csname urlstyle\endcsname\relax
  \providecommand{\doi}[1]{doi: #1}\else
  \providecommand{\doi}{doi: \begingroup \urlstyle{rm}\Url}\fi

\bibitem[Anil et~al.(2023)Anil, Dai, Firat, Johnson, Lepikhin, Passos, Shakeri, Taropa, Bailey, Chen, Chu, Clark, Shafey, Huang, Meier-Hellstern, Mishra, Moreira, Omernick, Robinson, Ruder, Tay, Xiao, Xu, Zhang, Abrego, Ahn, Austin, Barham, Botha, Bradbury, Brahma, Brooks, Catasta, Cheng, Cherry, Choquette-Choo, Chowdhery, Crepy, Dave, Dehghani, Dev, Devlin, Díaz, Du, Dyer, Feinberg, Feng, Fienber, Freitag, Garcia, Gehrmann, Gonzalez, Gur-Ari, Hand, Hashemi, Hou, Howland, Hu, Hui, Hurwitz, Isard, Ittycheriah, Jagielski, Jia, Kenealy, Krikun, Kudugunta, Lan, Lee, Lee, Li, Li, Li, Li, Li, Lim, Lin, Liu, Liu, Maggioni, Mahendru, Maynez, Misra, Moussalem, Nado, Nham, Ni, Nystrom, Parrish, Pellat, Polacek, Polozov, Pope, Qiao, Reif, Richter, Riley, Ros, Roy, Saeta, Samuel, Shelby, Slone, Smilkov, So, Sohn, Tokumine, Valter, Vasudevan, Vodrahalli, Wang, Wang, Wang, Wang, Wieting, Wu, Xu, Xu, Xue, Yin, Yu, Zhang, Zheng, Zheng, Zhou, Zhou, Petrov, and Wu]{anil2023palm}
Rohan Anil, Andrew~M. Dai, Orhan Firat, Melvin Johnson, Dmitry Lepikhin, Alexandre Passos, Siamak Shakeri, Emanuel Taropa, Paige Bailey, Zhifeng Chen, Eric Chu, Jonathan~H. Clark, Laurent~El Shafey, Yanping Huang, Kathy Meier-Hellstern, Gaurav Mishra, Erica Moreira, Mark Omernick, Kevin Robinson, Sebastian Ruder, Yi~Tay, Kefan Xiao, Yuanzhong Xu, Yujing Zhang, Gustavo~Hernandez Abrego, Junwhan Ahn, Jacob Austin, Paul Barham, Jan Botha, James Bradbury, Siddhartha Brahma, Kevin Brooks, Michele Catasta, Yong Cheng, Colin Cherry, Christopher~A. Choquette-Choo, Aakanksha Chowdhery, Clément Crepy, Shachi Dave, Mostafa Dehghani, Sunipa Dev, Jacob Devlin, Mark Díaz, Nan Du, Ethan Dyer, Vlad Feinberg, Fangxiaoyu Feng, Vlad Fienber, Markus Freitag, Xavier Garcia, Sebastian Gehrmann, Lucas Gonzalez, Guy Gur-Ari, Steven Hand, Hadi Hashemi, Le~Hou, Joshua Howland, Andrea Hu, Jeffrey Hui, Jeremy Hurwitz, Michael Isard, Abe Ittycheriah, Matthew Jagielski, Wenhao Jia, Kathleen Kenealy, Maxim Krikun, Sneha Kudugunta, Chang
  Lan, Katherine Lee, Benjamin Lee, Eric Li, Music Li, Wei Li, YaGuang Li, Jian Li, Hyeontaek Lim, Hanzhao Lin, Zhongtao Liu, Frederick Liu, Marcello Maggioni, Aroma Mahendru, Joshua Maynez, Vedant Misra, Maysam Moussalem, Zachary Nado, John Nham, Eric Ni, Andrew Nystrom, Alicia Parrish, Marie Pellat, Martin Polacek, Alex Polozov, Reiner Pope, Siyuan Qiao, Emily Reif, Bryan Richter, Parker Riley, Alex~Castro Ros, Aurko Roy, Brennan Saeta, Rajkumar Samuel, Renee Shelby, Ambrose Slone, Daniel Smilkov, David~R. So, Daniel Sohn, Simon Tokumine, Dasha Valter, Vijay Vasudevan, Kiran Vodrahalli, Xuezhi Wang, Pidong Wang, Zirui Wang, Tao Wang, John Wieting, Yuhuai Wu, Kelvin Xu, Yunhan Xu, Linting Xue, Pengcheng Yin, Jiahui Yu, Qiao Zhang, Steven Zheng, Ce~Zheng, Weikang Zhou, Denny Zhou, Slav Petrov, and Yonghui Wu.
\newblock Palm 2 technical report, 2023.

\bibitem[Bang(2023)]{bang2023gptcache}
Fu~Bang.
\newblock Gptcache: An open-source semantic cache for llm applications enabling faster answers and cost savings.
\newblock In \emph{Proceedings of the 3rd Workshop for Natural Language Processing Open Source Software (NLP-OSS 2023)}, pages 212--218, 2023.

\bibitem[Bast et~al.(2016)Bast, Buchhold, Haussmann, et~al.]{bast2016semantic}
Hannah Bast, Bj{\"o}rn Buchhold, Elmar Haussmann, et~al.
\newblock Semantic search on text and knowledge bases.
\newblock \emph{Foundations and Trends{\textregistered} in Information Retrieval}, 10\penalty0 (2-3):\penalty0 119--271, 2016.

\bibitem[Beeching et~al.(2023)Beeching, Belkada, Rasul, Tunstall, von Werra, Rajani, and Lambert]{beeching2023stackllama}
Edward Beeching, Younes Belkada, Kashif Rasul, Lewis Tunstall, Leandro von Werra, Nazneen Rajani, and Nathan Lambert.
\newblock Stackllama: An rl fine-tuned llama model for stack exchange question and answering, 2023.
\newblock URL \url{https://huggingface.co/blog/stackllama}.

\bibitem[Bommasani et~al.(2022)Bommasani, Hudson, Adeli, Altman, Arora, von Arx, Bernstein, Bohg, Bosselut, Brunskill, Brynjolfsson, Buch, Card, Castellon, Chatterji, Chen, Creel, Davis, Demszky, Donahue, Doumbouya, Durmus, Ermon, Etchemendy, Ethayarajh, Fei-Fei, Finn, Gale, Gillespie, Goel, Goodman, Grossman, Guha, Hashimoto, Henderson, Hewitt, Ho, Hong, Hsu, Huang, Icard, Jain, Jurafsky, Kalluri, Karamcheti, Keeling, Khani, Khattab, Koh, Krass, Krishna, Kuditipudi, Kumar, Ladhak, Lee, Lee, Leskovec, Levent, Li, Li, Ma, Malik, Manning, Mirchandani, Mitchell, Munyikwa, Nair, Narayan, Narayanan, Newman, Nie, Niebles, Nilforoshan, Nyarko, Ogut, Orr, Papadimitriou, Park, Piech, Portelance, Potts, Raghunathan, Reich, Ren, Rong, Roohani, Ruiz, Ryan, Ré, Sadigh, Sagawa, Santhanam, Shih, Srinivasan, Tamkin, Taori, Thomas, Tramèr, Wang, Wang, Wu, Wu, Wu, Xie, Yasunaga, You, Zaharia, Zhang, Zhang, Zhang, Zhang, Zheng, Zhou, and Liang]{bommasani2022opportunities}
Rishi Bommasani, Drew~A. Hudson, Ehsan Adeli, Russ Altman, Simran Arora, Sydney von Arx, Michael~S. Bernstein, Jeannette Bohg, Antoine Bosselut, Emma Brunskill, Erik Brynjolfsson, Shyamal Buch, Dallas Card, Rodrigo Castellon, Niladri Chatterji, Annie Chen, Kathleen Creel, Jared~Quincy Davis, Dora Demszky, Chris Donahue, Moussa Doumbouya, Esin Durmus, Stefano Ermon, John Etchemendy, Kawin Ethayarajh, Li~Fei-Fei, Chelsea Finn, Trevor Gale, Lauren Gillespie, Karan Goel, Noah Goodman, Shelby Grossman, Neel Guha, Tatsunori Hashimoto, Peter Henderson, John Hewitt, Daniel~E. Ho, Jenny Hong, Kyle Hsu, Jing Huang, Thomas Icard, Saahil Jain, Dan Jurafsky, Pratyusha Kalluri, Siddharth Karamcheti, Geoff Keeling, Fereshte Khani, Omar Khattab, Pang~Wei Koh, Mark Krass, Ranjay Krishna, Rohith Kuditipudi, Ananya Kumar, Faisal Ladhak, Mina Lee, Tony Lee, Jure Leskovec, Isabelle Levent, Xiang~Lisa Li, Xuechen Li, Tengyu Ma, Ali Malik, Christopher~D. Manning, Suvir Mirchandani, Eric Mitchell, Zanele Munyikwa, Suraj Nair,
  Avanika Narayan, Deepak Narayanan, Ben Newman, Allen Nie, Juan~Carlos Niebles, Hamed Nilforoshan, Julian Nyarko, Giray Ogut, Laurel Orr, Isabel Papadimitriou, Joon~Sung Park, Chris Piech, Eva Portelance, Christopher Potts, Aditi Raghunathan, Rob Reich, Hongyu Ren, Frieda Rong, Yusuf Roohani, Camilo Ruiz, Jack Ryan, Christopher Ré, Dorsa Sadigh, Shiori Sagawa, Keshav Santhanam, Andy Shih, Krishnan Srinivasan, Alex Tamkin, Rohan Taori, Armin~W. Thomas, Florian Tramèr, Rose~E. Wang, William Wang, Bohan Wu, Jiajun Wu, Yuhuai Wu, Sang~Michael Xie, Michihiro Yasunaga, Jiaxuan You, Matei Zaharia, Michael Zhang, Tianyi Zhang, Xikun Zhang, Yuhui Zhang, Lucia Zheng, Kaitlyn Zhou, and Percy Liang.
\newblock On the opportunities and risks of foundation models, 2022.

\bibitem[Borgeaud et~al.(2022)Borgeaud, Mensch, Hoffmann, Cai, Rutherford, Millican, Van Den~Driessche, Lespiau, Damoc, Clark, et~al.]{borgeaud2022improving}
Sebastian Borgeaud, Arthur Mensch, Jordan Hoffmann, Trevor Cai, Eliza Rutherford, Katie Millican, George~Bm Van Den~Driessche, Jean-Baptiste Lespiau, Bogdan Damoc, Aidan Clark, et~al.
\newblock Improving language models by retrieving from trillions of tokens.
\newblock In \emph{International conference on machine learning}, pages 2206--2240. PMLR, 2022.

\bibitem[Bubeck et~al.(2023)Bubeck, Chandrasekaran, Eldan, Gehrke, Horvitz, Kamar, Lee, Lee, Li, Lundberg, et~al.]{bubeck2023sparks}
S{\'e}bastien Bubeck, Varun Chandrasekaran, Ronen Eldan, Johannes Gehrke, Eric Horvitz, Ece Kamar, Peter Lee, Yin~Tat Lee, Yuanzhi Li, Scott Lundberg, et~al.
\newblock Sparks of artificial general intelligence: Early experiments with gpt-4.
\newblock \emph{arXiv preprint arXiv:2303.12712}, 2023.

\bibitem[Bura et~al.(2021)Bura, Rengarajan, Kalathil, Shakkottai, and Chamberland]{bura2021learning}
Archana Bura, Desik Rengarajan, Dileep Kalathil, Srinivas Shakkottai, and Jean-Francois Chamberland.
\newblock Learning to cache and caching to learn: Regret analysis of caching algorithms.
\newblock \emph{IEEE/ACM Transactions on Networking}, 30\penalty0 (1):\penalty0 18--31, 2021.

\bibitem[Chang et~al.(2020)Chang, Yu, Chang, Yang, and Kumar]{chang2020pre}
Wei-Cheng Chang, Felix~X Yu, Yin-Wen Chang, Yiming Yang, and Sanjiv Kumar.
\newblock Pre-training tasks for embedding-based large-scale retrieval.
\newblock \emph{arXiv preprint arXiv:2002.03932}, 2020.

\bibitem[Chang et~al.(2018)Chang, Lei, Zhou, Mao, and Ristaniemi]{chang2018learn}
Zheng Chang, Lei Lei, Zhenyu Zhou, Shiwen Mao, and Tapani Ristaniemi.
\newblock Learn to cache: Machine learning for network edge caching in the big data era.
\newblock \emph{IEEE Wireless Communications}, 25\penalty0 (3):\penalty0 28--35, 2018.

\bibitem[Chowdhery et~al.(2022)Chowdhery, Narang, Devlin, Bosma, Mishra, Roberts, Barham, Chung, Sutton, Gehrmann, et~al.]{chowdhery2022palm}
Aakanksha Chowdhery, Sharan Narang, Jacob Devlin, Maarten Bosma, Gaurav Mishra, Adam Roberts, Paul Barham, Hyung~Won Chung, Charles Sutton, Sebastian Gehrmann, et~al.
\newblock Palm: Scaling language modeling with pathways.
\newblock \emph{arXiv preprint arXiv:2204.02311}, 2022.

\bibitem[Faizal et~al.(2022)Faizal, Singh, Karamchandani, and Moharir]{faizal2022regret}
Fathima~Zarin Faizal, Priya Singh, Nikhil Karamchandani, and Sharayu Moharir.
\newblock Regret-optimal online caching for adversarial and stochastic arrivals.
\newblock In \emph{EAI International Conference on Performance Evaluation Methodologies and Tools}, pages 147--163. Springer, 2022.

\bibitem[Grave et~al.(2016)Grave, Joulin, and Usunier]{grave2016improving}
Edouard Grave, Armand Joulin, and Nicolas Usunier.
\newblock Improving neural language models with a continuous cache.
\newblock \emph{arXiv preprint arXiv:1612.04426}, 2016.

\bibitem[Grave et~al.(2017)Grave, Cisse, and Joulin]{grave2017unbounded}
Edouard Grave, Moustapha~M Cisse, and Armand Joulin.
\newblock Unbounded cache model for online language modeling with open vocabulary.
\newblock \emph{Advances in neural information processing systems}, 30, 2017.

\bibitem[He et~al.(2017)He, Zhang, Yu, Zhao, Yin, Leung, and Zhang]{he2017deep}
Ying He, Zheng Zhang, F~Richard Yu, Nan Zhao, Hongxi Yin, Victor~CM Leung, and Yanhua Zhang.
\newblock Deep-reinforcement-learning-based optimization for cache-enabled opportunistic interference alignment wireless networks.
\newblock \emph{IEEE Transactions on Vehicular Technology}, 66\penalty0 (11):\penalty0 10433--10445, 2017.

\bibitem[Izacard et~al.(2022)Izacard, Lewis, Lomeli, Hosseini, Petroni, Schick, Dwivedi-Yu, Joulin, Riedel, and Grave]{izacard2022few}
Gautier Izacard, Patrick Lewis, Maria Lomeli, Lucas Hosseini, Fabio Petroni, Timo Schick, Jane Dwivedi-Yu, Armand Joulin, Sebastian Riedel, and Edouard Grave.
\newblock Few-shot learning with retrieval augmented language models.
\newblock \emph{arXiv preprint arXiv:2208.03299}, 2022.

\bibitem[Jiang et~al.(2019)Jiang, Feng, Qin, Yum, and Cao]{jiang2019multi}
Wei Jiang, Gang Feng, Shuang Qin, Tak Shing~Peter Yum, and Guohong Cao.
\newblock Multi-agent reinforcement learning for efficient content caching in mobile d2d networks.
\newblock \emph{IEEE Transactions on Wireless Communications}, 18\penalty0 (3):\penalty0 1610--1622, 2019.

\bibitem[Johnson et~al.(2019)Johnson, Douze, and J{\'e}gou]{johnson2019billion}
Jeff Johnson, Matthijs Douze, and Herv{\'e} J{\'e}gou.
\newblock Billion-scale similarity search with {GPUs}.
\newblock \emph{IEEE Transactions on Big Data}, 7\penalty0 (3):\penalty0 535--547, 2019.

\bibitem[Kamalloo et~al.(2023)Kamalloo, Zhang, Ogundepo, Thakur, Alfonso-Hermelo, Rezagholizadeh, and Lin]{kamalloo2023evaluating}
Ehsan Kamalloo, Xinyu Zhang, Odunayo Ogundepo, Nandan Thakur, David Alfonso-Hermelo, Mehdi Rezagholizadeh, and Jimmy Lin.
\newblock Evaluating embedding apis for information retrieval.
\newblock \emph{arXiv preprint arXiv:2305.06300}, 2023.

\bibitem[Khandelwal et~al.(2019)Khandelwal, Levy, Jurafsky, Zettlemoyer, and Lewis]{khandelwal2019generalization}
Urvashi Khandelwal, Omer Levy, Dan Jurafsky, Luke Zettlemoyer, and Mike Lewis.
\newblock Generalization through memorization: Nearest neighbor language models.
\newblock \emph{arXiv preprint arXiv:1911.00172}, 2019.

\bibitem[Kumar and Singh(2016)]{kumar2016overview}
Swadhesh Kumar and PK~Singh.
\newblock An overview of modern cache memory and performance analysis of replacement policies.
\newblock In \emph{2016 IEEE International Conference on Engineering and Technology (ICETECH)}, pages 210--214. IEEE, 2016.

\bibitem[Kwiatkowski et~al.(2019)Kwiatkowski, Palomaki, Redfield, Collins, Parikh, Alberti, Epstein, Polosukhin, Kelcey, Devlin, Lee, Toutanova, Jones, Chang, Dai, Uszkoreit, Le, and Petrov]{47761}
Tom Kwiatkowski, Jennimaria Palomaki, Olivia Redfield, Michael Collins, Ankur Parikh, Chris Alberti, Danielle Epstein, Illia Polosukhin, Matthew Kelcey, Jacob Devlin, Kenton Lee, Kristina~N. Toutanova, Llion Jones, Ming-Wei Chang, Andrew Dai, Jakob Uszkoreit, Quoc Le, and Slav Petrov.
\newblock Natural questions: a benchmark for question answering research.
\newblock \emph{Transactions of the Association of Computational Linguistics}, 2019.

\bibitem[Lee et~al.(2001)Lee, Choi, Kim, Noh, Min, Cho, and Kim]{lee2001lrfu}
Donghee Lee, Jongmoo Choi, Jong-Hun Kim, Sam~H Noh, Sang~Lyul Min, Yookun Cho, and Chong~Sang Kim.
\newblock Lrfu: A spectrum of policies that subsumes the least recently used and least frequently used policies.
\newblock \emph{IEEE transactions on Computers}, 50\penalty0 (12):\penalty0 1352--1361, 2001.

\bibitem[Loshchilov and Hutter(2017)]{loshchilov2017decoupled}
Ilya Loshchilov and Frank Hutter.
\newblock Decoupled weight decay regularization.
\newblock \emph{arXiv preprint arXiv:1711.05101}, 2017.

\bibitem[Min et~al.(2022)Min, Shi, Lewis, Chen, Yih, Hajishirzi, and Zettlemoyer]{min2022nonparametric}
Sewon Min, Weijia Shi, Mike Lewis, Xilun Chen, Wen-tau Yih, Hannaneh Hajishirzi, and Luke Zettlemoyer.
\newblock Nonparametric masked language modeling.
\newblock \emph{arXiv preprint arXiv:2212.01349}, 2022.

\bibitem[Mukhopadhyay and Sinha(2021)]{mukhopadhyay2021online}
Samrat Mukhopadhyay and Abhishek Sinha.
\newblock Online caching with optimal switching regret.
\newblock In \emph{2021 IEEE International Symposium on Information Theory (ISIT)}, pages 1546--1551. IEEE, 2021.

\bibitem[Nori et~al.(2023)Nori, King, McKinney, Carignan, and Horvitz]{nori2023capabilities}
Harsha Nori, Nicholas King, Scott~Mayer McKinney, Dean Carignan, and Eric Horvitz.
\newblock Capabilities of gpt-4 on medical challenge problems.
\newblock \emph{arXiv preprint arXiv:2303.13375}, 2023.

\bibitem[OpenAI(2023)]{openai2023gpt4}
OpenAI.
\newblock Gpt-4 technical report, 2023.

\bibitem[Ouyang et~al.(2022)Ouyang, Wu, Jiang, Almeida, Wainwright, Mishkin, Zhang, Agarwal, Slama, Ray, et~al.]{ouyang2022training}
Long Ouyang, Jeffrey Wu, Xu~Jiang, Diogo Almeida, Carroll Wainwright, Pamela Mishkin, Chong Zhang, Sandhini Agarwal, Katarina Slama, Alex Ray, et~al.
\newblock Training language models to follow instructions with human feedback.
\newblock \emph{Advances in Neural Information Processing Systems}, 35:\penalty0 27730--27744, 2022.

\bibitem[Patterson et~al.(2021)Patterson, Gonzalez, Le, Liang, Munguia, Rothchild, So, Texier, and Dean]{patterson2021carbon}
David Patterson, Joseph Gonzalez, Quoc Le, Chen Liang, Lluis-Miquel Munguia, Daniel Rothchild, David So, Maud Texier, and Jeff Dean.
\newblock Carbon emissions and large neural network training.
\newblock \emph{arXiv preprint arXiv:2104.10350}, 2021.

\bibitem[Sharir et~al.(2020)Sharir, Peleg, and Shoham]{sharir2020cost}
Or~Sharir, Barak Peleg, and Yoav Shoham.
\newblock The cost of training nlp models: A concise overview.
\newblock \emph{arXiv preprint arXiv:2004.08900}, 2020.

\bibitem[Shuja et~al.(2021)Shuja, Bilal, Alasmary, Sinky, and Alanazi]{shuja2021applying}
Junaid Shuja, Kashif Bilal, Waleed Alasmary, Hassan Sinky, and Eisa Alanazi.
\newblock Applying machine learning techniques for caching in next-generation edge networks: A comprehensive survey.
\newblock \emph{Journal of Network and Computer Applications}, 181:\penalty0 103005, 2021.

\bibitem[Smith(1982)]{smith1982cache}
Alan~Jay Smith.
\newblock Cache memories.
\newblock \emph{ACM Computing Surveys (CSUR)}, 14\penalty0 (3):\penalty0 473--530, 1982.

\bibitem[Stallings(2011)]{stallings2011operating}
William Stallings.
\newblock \emph{Operating systems: internals and design principles}.
\newblock Prentice Hall Press, 2011.

\bibitem[Wang(1999)]{wang1999survey}
Jia Wang.
\newblock A survey of web caching schemes for the internet.
\newblock \emph{ACM SIGCOMM Computer Communication Review}, 29\penalty0 (5):\penalty0 36--46, 1999.

\bibitem[Wang et~al.(2022)Wang, Yang, Huang, Jiao, Yang, Jiang, Majumder, and Wei]{wang2022text}
Liang Wang, Nan Yang, Xiaolong Huang, Binxing Jiao, Linjun Yang, Daxin Jiang, Rangan Majumder, and Furu Wei.
\newblock Text embeddings by weakly-supervised contrastive pre-training.
\newblock \emph{arXiv preprint arXiv:2212.03533}, 2022.

\bibitem[Wang et~al.(2020)Wang, Salakhutdinov, and Yang]{wang_reinforcement_2020}
Ruosong Wang, Ruslan Salakhutdinov, and Lin~F. Yang.
\newblock Reinforcement {Learning} with {General} {Value} {Function} {Approximation}: {Provably} {Efficient} {Approach} via {Bounded} {Eluder} {Dimension}, 2020.
\newblock \_eprint: 2005.10804.

\bibitem[Wei et~al.(2022)Wei, Tay, Bommasani, Raffel, Zoph, Borgeaud, Yogatama, Bosma, Zhou, Metzler, et~al.]{wei2022emergent}
Jason Wei, Yi~Tay, Rishi Bommasani, Colin Raffel, Barret Zoph, Sebastian Borgeaud, Dani Yogatama, Maarten Bosma, Denny Zhou, Donald Metzler, et~al.
\newblock Emergent abilities of large language models.
\newblock \emph{arXiv preprint arXiv:2206.07682}, 2022.

\bibitem[Zhan et~al.(2022)Zhan, Huang, Huang, Jiang, and Lee]{zhan2022offline}
Wenhao Zhan, Baihe Huang, Audrey Huang, Nan Jiang, and Jason Lee.
\newblock Offline reinforcement learning with realizability and single-policy concentrability.
\newblock In \emph{Conference on Learning Theory}, pages 2730--2775. PMLR, 2022.

\bibitem[Zhong et~al.(2022)Zhong, Lei, and Chen]{zhong2022training}
Zexuan Zhong, Tao Lei, and Danqi Chen.
\newblock Training language models with memory augmentation.
\newblock \emph{arXiv preprint arXiv:2205.12674}, 2022.

\bibitem[Zhu et~al.(2023{\natexlab{a}})Zhu, Sheng, Zheng, Barrett, Jordan, and Jiao]{zhu2023optimal}
Banghua Zhu, Ying Sheng, Lianmin Zheng, Clark Barrett, Michael~I Jordan, and Jiantao Jiao.
\newblock On optimal caching and model multiplexing for large model inference.
\newblock \emph{arXiv preprint arXiv:2306.02003}, 2023{\natexlab{a}}.

\bibitem[Zhu and Zhang(2023)]{zhu2023provably}
Hanlin Zhu and Amy Zhang.
\newblock Provably efficient offline goal-conditioned reinforcement learning with general function approximation and single-policy concentrability.
\newblock \emph{arXiv preprint arXiv:2302.03770}, 2023.

\bibitem[Zhu et~al.(2023{\natexlab{b}})Zhu, Wang, and Lee]{zhu_provably_2023}
Hanlin Zhu, Ruosong Wang, and Jason Lee.
\newblock Provably {Efficient} {Reinforcement} {Learning} via {Surprise} {Bound}.
\newblock In \emph{International {Conference} on {Artificial} {Intelligence} and {Statistics}}, pages 4006--4032. PMLR, 2023{\natexlab{b}}.

\bibitem[Ziegler et~al.(2019)Ziegler, Stiennon, Wu, Brown, Radford, Amodei, Christiano, and Irving]{ziegler2019fine}
Daniel~M Ziegler, Nisan Stiennon, Jeffrey Wu, Tom~B Brown, Alec Radford, Dario Amodei, Paul Christiano, and Geoffrey Irving.
\newblock Fine-tuning language models from human preferences.
\newblock \emph{arXiv preprint arXiv:1909.08593}, 2019.

\bibitem[zilliztech(2023)]{zilliztech_2023}
zilliztech.
\newblock Gptcache: Semantic cache for llms. fully integrated with langchain and llama$\_$index., 2023.
\newblock URL \url{https://github.com/zilliztech/GPTCache}.

\end{thebibliography}
